\documentclass{article} % For LaTeX2e
\usepackage{iclr2027_conference,times}

\usepackage{amsmath,amsfonts,bm}

\def\eqref#1{equation~\ref{#1}}
\def\1{\bm{1}}

\DeclareMathAlphabet{\mathsfit}{\encodingdefault}{\sfdefault}{m}{sl}
\SetMathAlphabet{\mathsfit}{bold}{\encodingdefault}{\sfdefault}{bx}{n}

\usepackage{hyperref}
\usepackage{url}
\usepackage{graphicx}
\usepackage{booktabs}
\usepackage{multirow}
\usepackage{colortbl}
\usepackage{algorithm}
\usepackage{algpseudocode}
\usepackage{enumitem}
\usepackage{amssymb}

\usepackage{xspace}
\definecolor{crblue}{RGB}{232,240,248}
\newcommand{\sysname}{\textsc{ContextRender}\xspace}
\title{ContextRender: From Execution Dependencies to Agent Context}

\author{Savini Kashmira, Jayanaka L. Dantanarayana, Lingjia Tang \& Jason Mars \\
University of Michigan, Ann Arbor, USA \\
\texttt{\{savinik,jayanaka,lingjia,profmars\}@umich.edu}
}

\iclrfinalcopy
\begin{document}

\maketitle

% arXiv version: no ICLR running header. Delete these two lines for the
% ICLR submission, where the header is part of the required format.
\lhead{}
\renewcommand{\headrulewidth}{0pt}

\begin{abstract}
LLM agents performing long-horizon tasks accumulate tool results that
later steps may need. Passing the full history to every invocation is
costly even when it fits within the context window, while reducing it
risks omitting needed information. Existing context management methods
can overlook how earlier tool results are used in subsequent execution,
leaving needed information out of context.
We introduce \sysname{}, which manages context through a persistent
graph of execution dependencies. We develop Tool-Flow Analysis
to track how later operations reuse information from earlier tool results,
providing a signal called \emph{observed reuse}. A renderer combines
this signal with recency and semantic relevance to select results within
a fixed history budget, retaining omitted results for later use.
Across AppWorld and 8-objective QA with three execution models,
\sysname{} outperforms the evaluated context management baselines
using a 6K history budget, well below the models' maximum context
windows. Within this budget, it achieves task performance close to
or above that of passing the full history while reducing mean inference cost by 10.2\%--32.2\% relative to
\textit{Full history}. Ablations show that observed reuse improves
task performance and retention of results reused later.
\end{abstract}

\section{Introduction}
\label{sec:intro}

Large Language Model (LLM) agents increasingly use tools to carry out complex, long-horizon tasks such as software engineering, deep research, and work across digital applications \citep{yang2024sweagent,trivedi2024appworld}. Executing these tasks often requires an agent to make tens to hundreds of model calls and tool interactions, causing its interaction history to grow throughout execution. Information accumulated in this history may be needed again as the agent carries out later steps. However, carrying the entire accumulated interaction history forward at every invocation imposes substantial inference cost, can exceed the context window, and can expose the model to irrelevant context, adding noise to the current step \citep{liu2024lostinthemiddle,zhang2026masking}. Managing this history can reduce inference cost even when it fits within the model's context window.

\begin{figure*}[t]
    \centering
    \includegraphics[width=\linewidth]{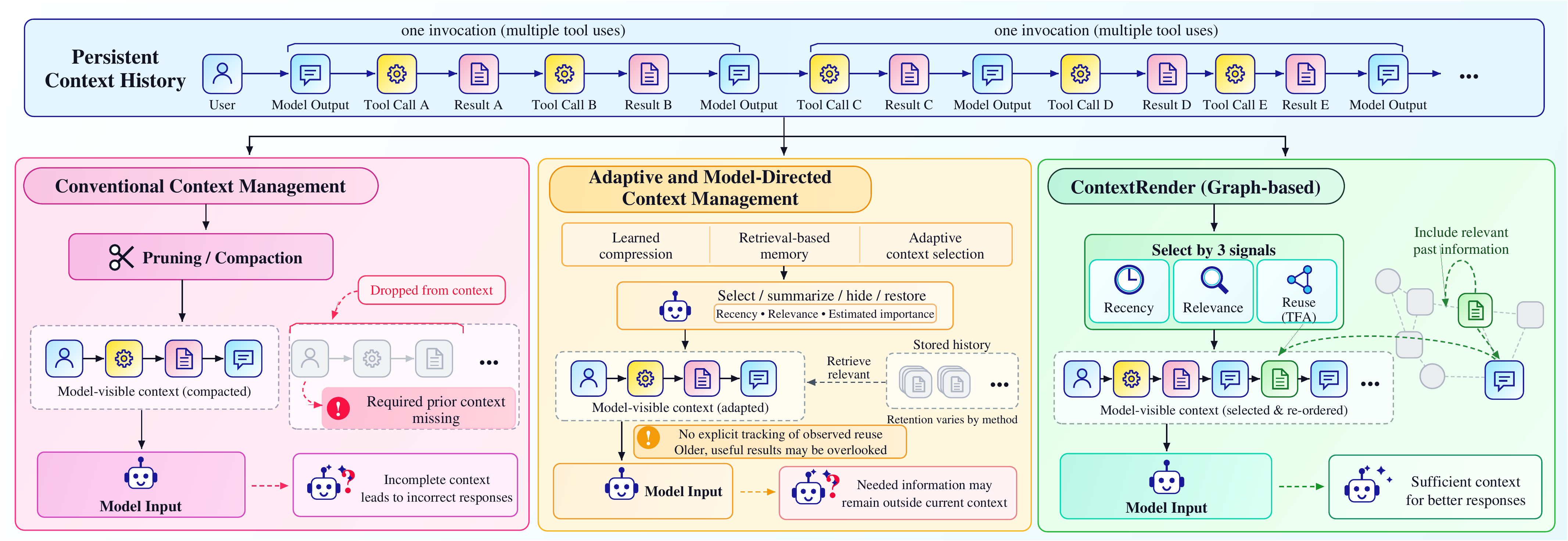}
    % \vspace{-0.5cm}
\caption{Comparison of context management approaches.
\sysname{} selects history from a persistent graph using recency,
semantic relevance, and observed reuse, retaining unselected
results for later use.}
    \label{fig:overview}
    \vspace{-0.5cm}
\end{figure*}

To limit context growth, agent systems such as OpenClaw, OpenCode, and Hermes Agent provide mechanisms for managing accumulated interaction history \citep{openclaw2026,opencode2025,hermes2026}. When configured context-size thresholds are reached, the model-visible history can be reduced using techniques such as pruning and compaction. Pruning removes selected messages or tool results based on heuristics such as recency or size, while compaction replaces earlier interactions with a shorter summary \citep{kang2025acon,lindenbauer2025complexitytrap,sun2025contextfolding,ye2025agentfold,dang2026arc}. However, these operations may leave details from earlier tool results absent or incomplete in the model-visible context when needed later in execution (Figure~\ref{fig:overview}, left). If the complete tool results are not retained elsewhere, recovering those details may require additional tool calls \citep{lindenbauer2025complexitytrap,zhang2026masking}.

More recent work has sought to preserve excluded information and improve context selection. Learned and optimized compression attempts to retain important details when earlier interactions are shortened \citep{kang2025acon,sun2025contextfolding,ye2025agentfold}, while retrieval-based memory preserves earlier information outside the model-visible context for later access \citep{packer2023memgpt}. Adaptive and model-directed approaches adjust the visible context during execution. ACE selects raw, abstracted, or hidden historical steps based on the current task state \citep{liao2026ace}, while PACE adjusts historical granularity according to predicted relevance to the next action \citep{wei2026pace}. Sculptor provides model-controlled context editing and restoration \citep{li2025sculptor}, and Self-GC uses a planner to propose folding, masking, and pruning operations \citep{hao2026selfgc} (Figure~\ref{fig:overview}, middle). However, judging which information to expose remains difficult when an earlier result supports later operations in ways that are not apparent from its relevance to the current step. A result containing repeatedly used identifiers or other values may receive low priority even while those values remain useful. This motivates supplementing relevance judgments with evidence of how earlier results have been used during execution, helping identify which historical tool results to show at each invocation.

Our key insight is that \textbf{context management should track execution dependencies rather than just maintain a chronological list of interactions, and use these dependencies to guide which earlier tool results are shown to the model at each invocation}. These dependencies connect earlier tool results to later model steps and tool calls that reuse values those results introduced. Tracking these relationships captures how results are used as execution progresses, beyond when they were produced or how closely their content matches the current invocation. We call this evidence of use \textit{observed reuse}. An earlier tool result may remain useful because subsequent operations continue to draw on its information, even when recency or semantic relevance gives it low priority. Context selection can therefore combine this evidence with recency and semantic relevance to retain or restore complete tool results at later invocations.

Building on this insight, we introduce \textbf{\sysname{}}, a context-management system that records execution history and detected reuse relationships in a persistent context graph, then selects which earlier tool results to show at each model invocation under a fixed history budget (Figure~\ref{fig:overview}, right). \sysname{} has three components: (1) The persistent context graph stores messages, tool calls, and complete tool results while preserving their chronological order. (2) Tool-Flow Analysis (TFA) tracks how later model steps and tool calls reuse values from earlier tool results. It links each detected use to the tool result that first introduced the value and records this evidence as \textit{observed reuse}. (3) A renderer combines observed reuse with recency and semantic relevance to select complete historical tool results within the allocated budget. Unselected results remain available in the graph for later invocations. Figure~\ref{fig:contextrender} illustrates the overall design.

We compare \sysname{} with baseline context-management methods using three execution models. Our evaluation shows that \sysname{} can achieve task performance close to or above passing the full accumulated history using a small history budget for including the agent's previous execution steps in context, well below the models' maximum context windows. At our primary 6K setting, \sysname{} consistently achieves the highest task performance among the evaluated managed-context baselines across two benchmarks, AppWorld \citep{trivedi2024appworld} and 8-objective QA~\citep{kwiatkowski2019nq,zhou2025mem1}, with all three models. It achieves this performance while reducing mean standardized inference cost by 10.2\%--32.2\% relative to passing the full history. Our ablation study shows that adding observed reuse from recorded execution dependencies to recency and semantic relevance improves task performance and retention of results reused later, with particularly large recall gains for results reused after long intervals.

This paper makes the following contributions.
\begin{enumerate}[nosep]
     \item We introduce Tool-Flow Analysis (TFA), which tracks how later operations reuse information from earlier tool results, providing an observed-reuse signal for context selection.
    \item We introduce \sysname{}, which uses TFA and a persistent context graph to include earlier tool results in the model's context within a token budget allocated to history, while retaining omitted results for later use.
    \item We evaluate \sysname{} across two benchmarks with three execution models, demonstrating comparable or higher task performance while reducing inference cost by approximately 10\%--32\% relative to passing the full accumulated history.
\end{enumerate}
\section{Motivation}

\begin{figure}[t]
\centering
\includegraphics[width=0.8\linewidth]{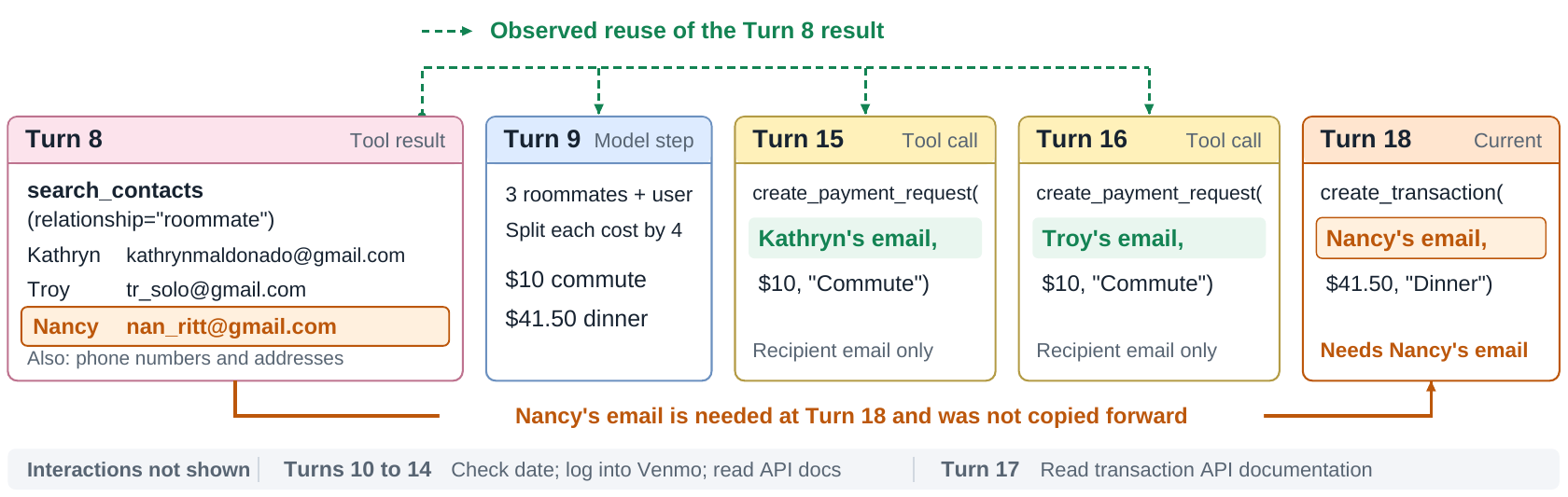}
\caption{Execution trace from an AppWorld task in which the agent settles a shared dinner bill among roommates. Information from the contact result produced at Turn 8 is used by later operations at Turns 9, 15, and 16. At Turn 18, the agent needs Nancy's email address, which has not been copied forward and remains available only from the Turn 8 result.}
\label{fig}
\end{figure}

AI agents need access to earlier tool results to complete long-horizon tasks successfully, but passing the full history at every invocation incurs inference cost. Selecting earlier tool results by recency or task relevance within a history token budget for previous execution steps can reduce cost but risks omitting information needed later. We illustrate this using an AppWorld task in which an agent is asked to settle a shared dinner bill among roommates on Venmo (Figure~\ref{fig}).

\subsection{Limitations of Recency and Relevance}

At Turn 8, the agent calls \texttt{search\_contacts(relationship="roommate")}, which returns contact information for Kathryn, Troy, and Nancy. At Turn 9, it uses the three names to determine that the bill should be split four ways. At Turns 15 and 16, it uses Kathryn's and Troy's email addresses to create payment requests. Each call contains only its recipient's email address, without copying the full contact list. At Turn 18, the agent needs Nancy's email address to send her the dinner payment. Her email has not been carried forward and remains available only from the Turn 8 result. Figure~\ref{fig} illustrates these execution dependencies between the Turn 8 result and the later operations that use its information.

The execution log shows that, when selecting earlier tool results within the history budget, both recency and task relevance omitted the Turn 8 result at Turn 18. Recency favors results from later turns, leaving the Turn 8 result outside the retained history. Task relevance favors results about Venmo APIs and payment operations because their content more closely matches the current payment step. Either criterion can therefore leave information needed for the current step outside the model-visible context.

However, the payment calls at Turns 15 and 16 show that information
from the Turn 8 result continues to be used. This past use provides
additional evidence that the result may remain useful, beyond its
age or similarity to the current step. Retaining the complete
result also preserves Nancy's email, even though her address has
not yet been reused. This motivates considering a result's previous
use when deciding whether to retain it for later steps.

\subsection{Execution Dependencies for Context Selection}

Using observed reuse to guide context selection requires recording \emph{execution dependencies} between earlier tool results and the later operations that reuse their information. A chronological history records when the tool result appeared and when subsequent operations occurred. A graph-structured representation can explicitly record these dependencies by linking earlier tool results to later operations that reuse their information, making the detected reuse available for context selection.

Observed reuse provides additional evidence beyond recency and task relevance for selecting earlier tool results that may be needed later. The goal is to preserve this information within a small history budget, supporting performance close to or above passing the full history. \sysname{} implements this approach using a persistent context graph and Tool-Flow Analysis to guide context selection.

\section{\sysname}
\label{sec:method}

\begin{figure}[t]
\centering
\includegraphics[width=0.9\linewidth]{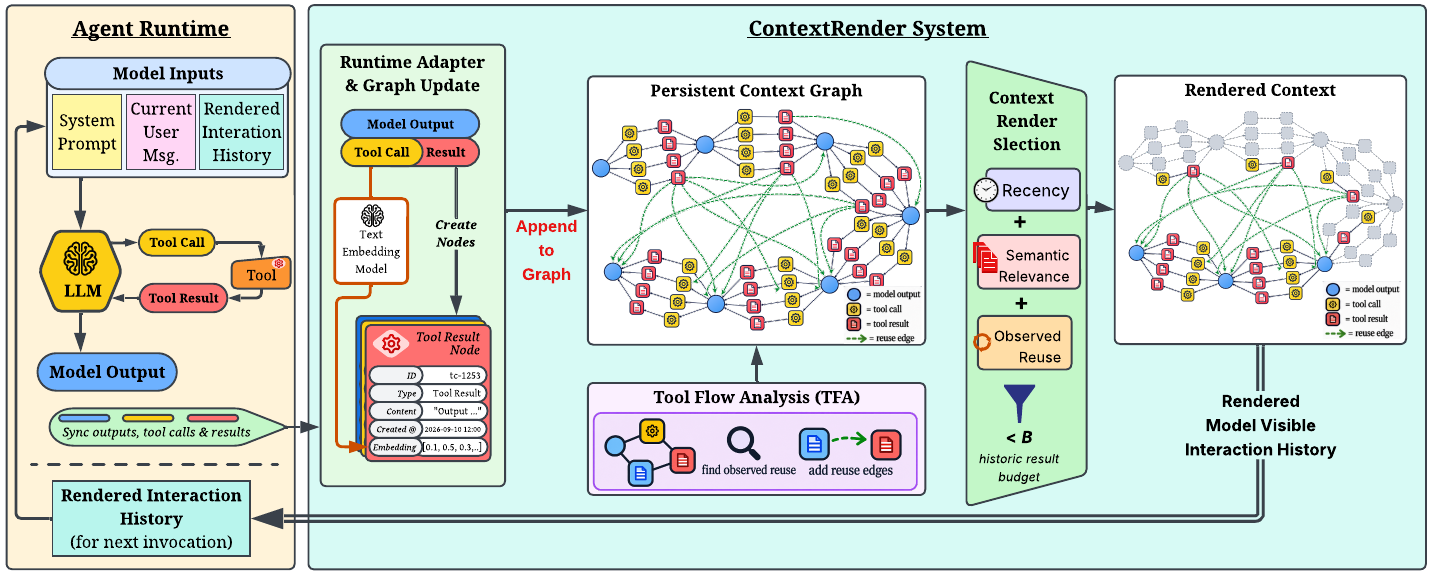}
\caption{\sysname architecture. The runtime adapter records messages,
tool calls, and complete tool results in a persistent context graph.
Tool-Flow Analysis records observed reuse as execution dependencies.
Before each model invocation, the renderer includes the accumulated
execution history if it fits within the history budget $B$; otherwise,
it selects tool results using observed reuse, recency, and semantic
relevance. Omitted results remain stored for later invocations.}
\label{fig:contextrender}
\end{figure}

At each model invocation, \sysname selects earlier tool results from
a persistent context graph under a fixed history budget $B$. As shown
in Figure~\ref{fig:contextrender}, the system has three components:
(1) a persistent context graph that stores messages, tool calls, and
complete tool results; (2) Tool-Flow Analysis (TFA), which records
observed reuse as execution dependencies by linking later uses of
values to earlier tool results; and (3) a renderer that combines this
observed reuse with recency and semantic relevance to select results
within the available budget.

\subsection{Persistent Context Graph}
\label{sec:context-state}

During task execution, \sysname maintains a persistent context
graph $G$ containing messages, tool calls, and tool results in
chronological order, together with the reuse edges identified
by TFA (Section~\ref{sec:usefulness}). Each tool result $r$ is
stored with its complete content, position, timestamp, and
content embedding. Result age and content embeddings support
recency and semantic relevance, respectively.

\subsection{Tool-Flow Analysis}
\label{sec:usefulness}

TFA tracks how values from earlier tool results are reused in
later model outputs and tool calls. It identifies qualifying
values using the extraction rules in Appendix~\ref{app:reference-matching} and detects
reuse through exact, case-sensitive matching. Each detected use
is linked to the first earlier tool result containing that value;
intermediate copies do not change this attribution. We call this
evidence of use \emph{observed reuse} and record the corresponding
links as reuse edges in the persistent context graph
(Figure~\ref{fig:contextrender}).

For example, suppose a tool result introduces
\texttt{customer\_id=C142XYZ}. If a later model output repeats
\texttt{C142XYZ} and a subsequent tool call uses it, TFA links both
uses to the tool result that first introduced the value. Because
information from this result is reused in multiple later steps,
these uses increase its reuse score and give it higher priority
for inclusion in context.

TFA summarizes this evidence with a reuse score $c_t(r)$. Let
$\mathcal{M}_t(r)$ denote the detected reuse matches observed before
invocation $t$ and attributed to result $r$, counting each distinct
value once per later model output or tool call. We define
\begin{equation}
c_t(r)
=
\sum_{m \in \mathcal{M}_t(r)}
\frac{1}{n_t(m)},
\label{eq:reuse-mass}
\end{equation}
where $n_t(m)$ is the number of historical tool results in the graph
before invocation $t$ that contain the matched value. This weighting
reduces the contribution of values shared by multiple tool results.

\subsection{Context Rendering}
\label{sec:rendering}

Before each model invocation, the renderer checks whether the
agent's accumulated execution history fits within the history
budget $B$, which limits tokens for information from previous
execution steps. If the history fits within $B$, it is included
in the model input. Otherwise, the renderer selects complete
historical tool results using the scores below to construct
context within the available history budget.
Appendix~\ref{app:budget-accounting} specifies the token estimate
and budget accounting.

Let $a_t(r)$ be the number of turns since result $r$ was produced,
where one turn corresponds to one model invocation. Recency is
represented by $e^{-\lambda a_t(r)}$, where $\lambda$ controls its
decay. For semantic relevance, let $x$ denote the task's first user
message and $f_t$ the latest visible interaction before invocation
$t$, formed by concatenating the latest visible user and assistant
messages. We construct
\[
q_t
=
0.4\,\operatorname{unit}(\operatorname{emb}(x))
+
0.6\,\operatorname{unit}(\operatorname{emb}(f_t)),
\]
where $\operatorname{unit}(\cdot)$ denotes vector normalization. We
define the relevance score as
\[
s_t(r)
=
\max\!\left(
\cos(\operatorname{emb}(r),q_t),0
\right).
\]
Observed reuse contributes through $1-e^{-\gamma c_t(r)}$. This
function gives additional reuse evidence a diminishing contribution,
with $\gamma$ controlling how quickly the term saturates.

The resulting usefulness estimate is
\begin{equation}
U_t(r)
=
W_{\mathrm{rec}} e^{-\lambda a_t(r)}
+
W_{\mathrm{reuse}}\left(1-e^{-\gamma c_t(r)}\right)
+
W_{\mathrm{rel}} s_t(r),
\label{eq:usefulness}
\end{equation}
where $W_{\mathrm{rec}}$, $W_{\mathrm{reuse}}$, and
$W_{\mathrm{rel}}$ weight the three signals. Recency favors recently
produced results, semantic relevance favors results related to the
task and latest interaction, and observed reuse contributes evidence
from later operations that reused their values. Appendix~A.3 gives
the embedding configuration and parameter values.

After computing $U_t(r)$ for each historical tool result, the
renderer selects results within the available history budget.
It prioritizes results with higher usefulness scores while
avoiding the selection of multiple results with very similar
content. Each selected result is included in the model input
with its complete content from the graph. Unselected results
remain in the persistent graph. Before subsequent model
invocations, usefulness scores are recomputed from the updated
graph, allowing previously omitted results to be selected as
selection priorities change. Appendix~\ref{app:budget-accounting}
gives the exact selection and budget-accounting procedure.

\subsection{Runtime Integration}
\label{sec:runtime}

As illustrated in Figure~\ref{fig:contextrender}, \sysname runs as a
separate server connected to the agent runtime through an adapter.
The adapter sends new messages, tool calls, and tool results to the
server, which adds them to the persistent graph and updates the
reuse relationships identified by TFA. Before each model invocation,
the server returns the context rendered under budget $B$ to the
adapter, which supplies it to the model. The agent otherwise
continues its normal message and tool execution.

We implement \sysname in Jac~\citep{mars2023jaseci}, using its Object
Spatial Programming abstractions~\citep{mars2025osp} to represent and
operate over the persistent context graph.

\providecommand{\sysname}{ContextRender\xspace}
\definecolor{crblue}{RGB}{230,240,250}
\section{Evaluation}
\label{sec:evaluation}

We evaluate how \sysname affects task performance and inference
cost compared with passing the full execution history without
context management and with the selected context management
baselines. We also conduct an ablation study to evaluate the
contribution of observed reuse and examine how task performance
varies with the history budget.

\subsection{Benchmarks and Baselines}
\label{sec:eval-benchmarks}

\paragraph{Benchmarks.}
We evaluate on two long-horizon agentic benchmarks.
\textbf{AppWorld}~\citep{trivedi2024appworld} involves multi-step tool use
across 9 simulated applications and approximately 100 users. We use
all 168 Test-Normal tasks and all 417 Test-Challenge tasks with
AppWorld's ReAct agent.
\textbf{8-objective QA}~\citep{kwiatkowski2019nq,zhou2025mem1} requires
answering 8 NaturalQuestions queries using a search tool and
returning a consolidated answer set. We evaluate 100 test tasks
using ACON's released agent implementation~\citep{kang2025acon}.

\paragraph{Baselines.}
\label{sec:eval-baselines}
Motivated by prior work on compact agent
histories~\citep{kang2025acon,yang2024sweagent}, we use a history
budget $B$ for including the agent's previous execution steps
in context. We set $B=6$K in the primary setting.
We compare \sysname with 5 baselines, all
constrained by the same budget $B$:
\textbf{Prune} retains the most recent results that fit within
$B$~\citep{yang2024sweagent,lindenbauer2025complexitytrap};
\textbf{Compact (LLM)} uses an LLM to compress the entire
interaction history to fit within
$B$~\citep{lindenbauer2025complexitytrap};
\textbf{Recency + Summary} retains recent results and summarizes
older ones, fitting both within
$B$~\citep{smith2025openhands,packer2023memgpt};
\textbf{Semantic Retrieval} selects results within $B$ by embedding
similarity to the latest user and assistant
messages~\citep{xu2025amem}; and
\textbf{ACON} uses the released implementation with the authors'
optimized compression guidelines under $B$~\citep{kang2025acon}.
We also include \textbf{Full history}, which passes the entire
accumulated execution history without context management even when
it exceeds $B$, as a reference for comparing task performance and
inference cost.

\begin{figure*}[t]
\centering
\begin{minipage}[t]{0.325\textwidth}
    \centering
    \includegraphics[width=\linewidth]{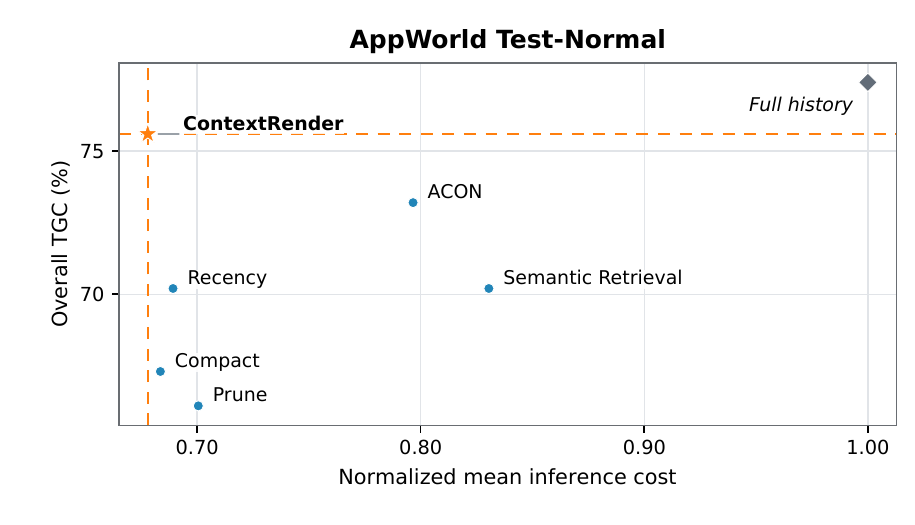}\\[0pt]
    {\small (a) GPT-4.1}
\end{minipage}%
\hfill
\begin{minipage}[t]{0.325\textwidth}
    \centering
    \includegraphics[width=\linewidth]{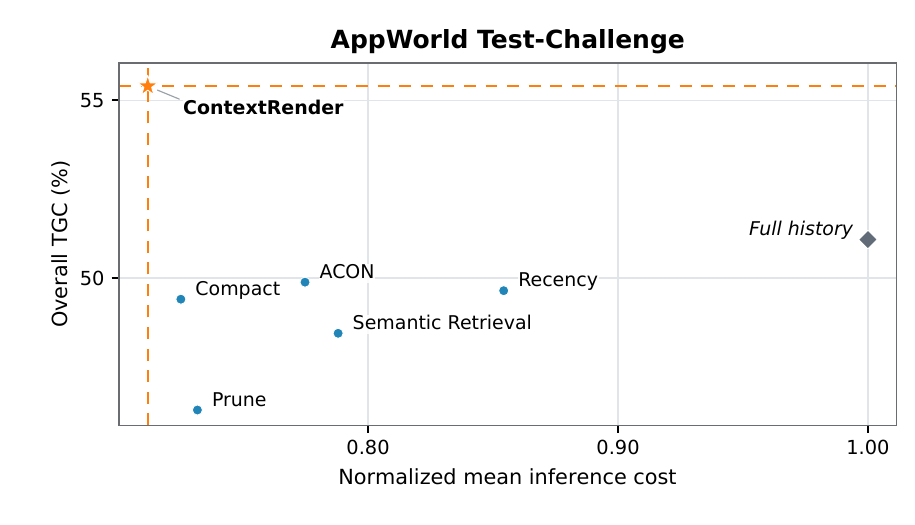}\\[0pt]
    {\small (b) GPT-4.1}
\end{minipage}%
\hfill
\begin{minipage}[t]{0.325\textwidth}
    \centering
    \includegraphics[width=\linewidth]{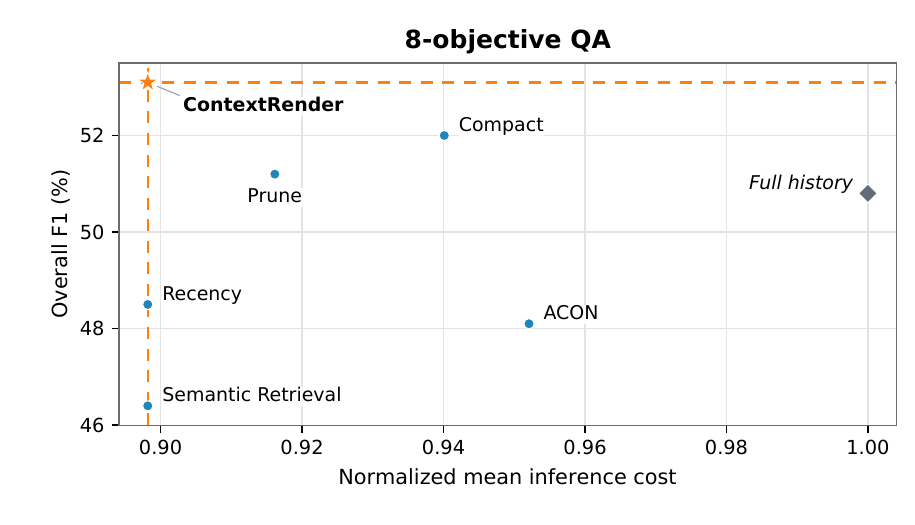}\\[0pt]
    {\small (c) GPT-4.1}
\end{minipage}%
\par\smallskip
\begin{minipage}[t]{0.325\textwidth}
    \centering
    \includegraphics[width=\linewidth]{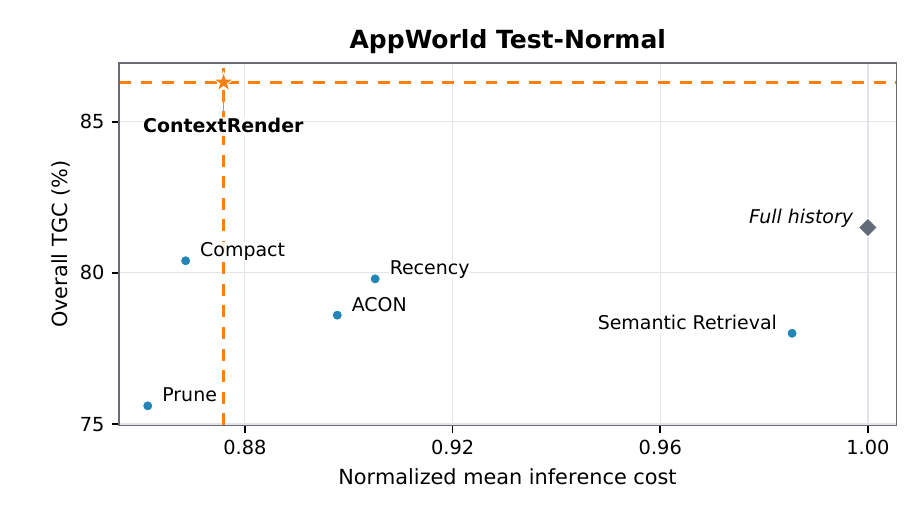}\\[0pt]
    {\small (d) Muse Glimmer 30B}
\end{minipage}%
\hfill
\begin{minipage}[t]{0.325\textwidth}
    \centering
    \includegraphics[width=\linewidth]{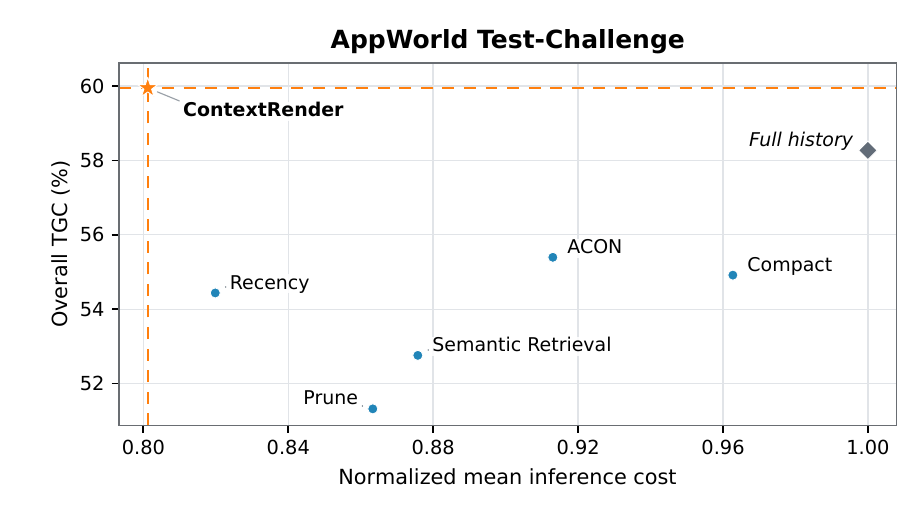}\\[0pt]
    {\small (e) Muse Glimmer 30B}
\end{minipage}%
\hfill
\begin{minipage}[t]{0.325\textwidth}
    \centering
    \includegraphics[width=\linewidth]{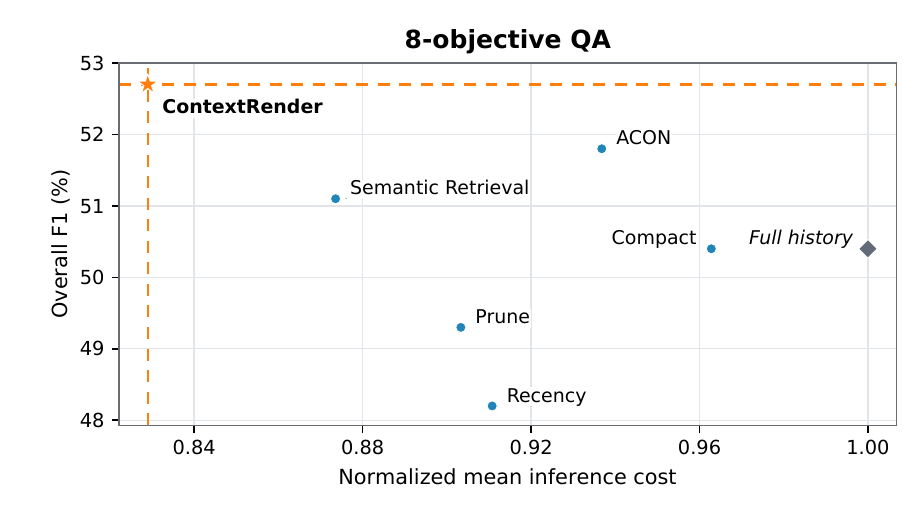}\\[0pt]
    {\small (f) Muse Glimmer 30B}
\end{minipage}%
\par\smallskip
\begin{minipage}[t]{0.325\textwidth}
    \centering
    \includegraphics[width=\linewidth]{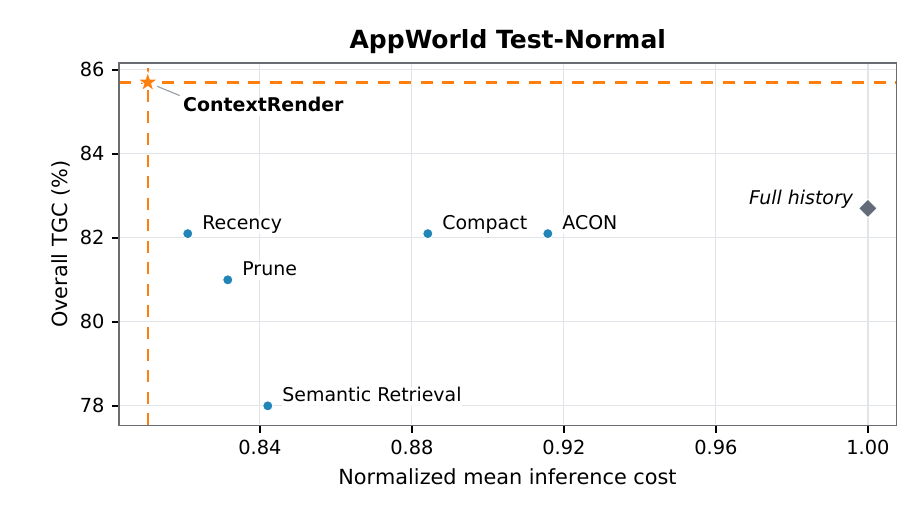}\\[0pt]
    {\small (g) DeepSeek V4 Pro}
\end{minipage}%
\hfill
\begin{minipage}[t]{0.325\textwidth}
    \centering
    \includegraphics[width=\linewidth]{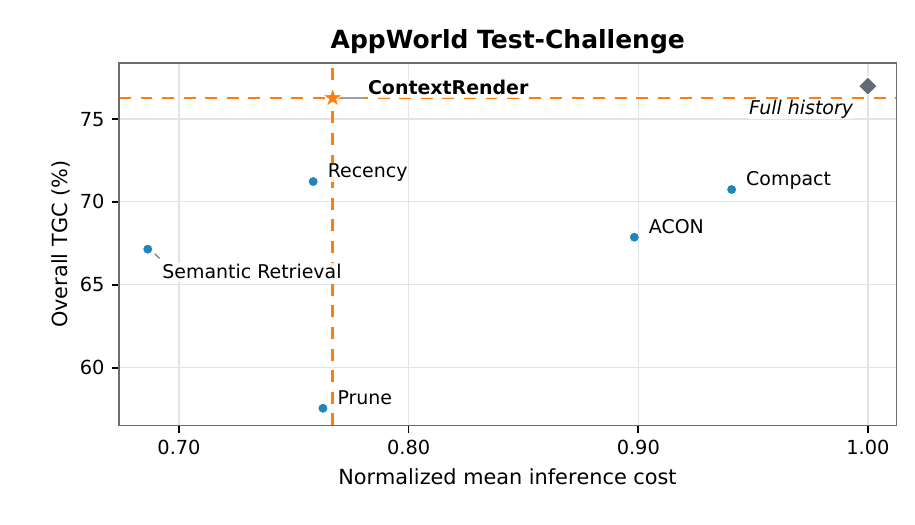}\\[0pt]
    {\small (h) DeepSeek V4 Pro}
\end{minipage}%
\hfill
\begin{minipage}[t]{0.325\textwidth}
    \centering
    \includegraphics[width=\linewidth]{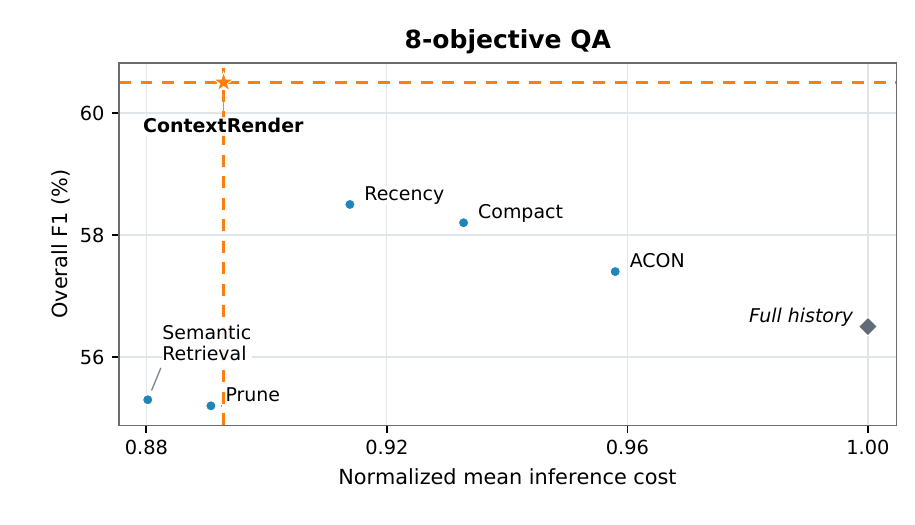}\\[0pt]
    {\small (i) DeepSeek V4 Pro}
\end{minipage}%
\par\smallskip
\caption{Overall performance (AppWorld TGC; QA F1) across all tasks
versus mean standardized inference cost on the same tasks that
trigger context management at $B=6$K. Costs are normalized to the
mean cost of \textit{Full history} ($=1$).
``Recency'' denotes Recency + Summary.}
\label{fig:rq3-cost}
\end{figure*}

\subsection{Evaluation Methodology}
\label{sec:eval-methodology}

\paragraph{Models and budgets.}
We evaluate GPT-4.1, Muse Glimmer 30B, and DeepSeek V4 Pro.
With GPT-4.1, we sweep $B\in\{3,6,12,24\}$K on AppWorld Test-Normal
and $B\in\{2,4,6,8,12\}$K on 8-objective QA.
For each benchmark and model, all compared systems use the same
agent prompt, tool interface, and execution limits.

\paragraph{Metrics.}
We report AppWorld Task Goal Completion (TGC, \%), the percentage of tasks
passing the benchmark's task-goal checks, and token-level F1 (\%) on
8-objective QA. These primary performance metrics are evaluated over
all tasks. \textbf{Later-Used Result Recall} measures how often the
complete earlier tool result is available in the model's context when
its information is reused, expressed as a percentage.

\paragraph{Inference cost.}
Inference cost includes all model and embedding calls.
We use the following cost model
to price token usage consistently across execution models,
instead of using provider-specific prices. Token prices are
expressed in USD per million tokens:
\begin{equation}
C=\frac{[2(1-\rho)+0.5\rho]P+8O+0.13E}{10^6},
\label{eq:standardized-inference-cost}
\end{equation}
where $P$, $O$, and $E$ are the total input, output, and embedding
token counts for a task, and $\rho$ is the fraction of
input tokens served from cache.
We use the cache rate observed in \textit{Full history} to price
all methods evaluated with the same model on the same benchmark.
Appendix~\ref{app:cost-standardization} details token accounting
and normalization.

% \paragraph{Ablation protocol.}
% To isolate each selection signal in Section~\ref{sec:rq2}, we compare
% four \sysname variants: \textbf{CR--Recency} uses result age,
% \textbf{CR--Relevance} uses semantic relevance,
% \textbf{CR--Recency + Relevance} combines both, and
% \textbf{CR--Full (+Reuse)} adds observed reuse.
% We evaluate GPT-4.1 on AppWorld Test-Normal and QA at $B=6$K,
% keeping all other components fixed.
% To measure Later-Used Result Recall, we first record GPT-4.1 executions
% using the \textit{Full history} baseline. Before each model invocation
% in these recordings, we apply each variant's selection rule to the
% history available at that point to determine which tool results it
% would include.
% We identify reuse when a later assistant turn contains an exact match
% to a token first introduced by an earlier tool result. Appendix~\ref{app:recall-protocol} provides detailed rules.

\subsection{Task Performance and Inference Efficiency}
\label{sec:rq1}
\label{sec:rq3}

We measure overall task performance across all tasks in AppWorld
Test-Normal, Test-Challenge, and 8-objective QA for each execution
model. To assess inference efficiency, we measure mean inference
cost across methods on the same tasks that trigger context
management when their accumulated execution history exceeds the
6K history budget. We also evaluate \textit{Full history} as a
reference for task performance and inference cost when the entire
accumulated execution history is passed without context management.

Figure~\ref{fig:rq3-cost} plots overall task performance against
mean standardized inference cost, normalized to that of
\textit{Full history} on the same tasks in each model--benchmark
setting (\textit{Full history} $=1$).
Table~\ref{tab:relative-performance-cost} reports the corresponding
relative performance and cost changes of \sysname against each
baseline and the \textit{Full history} reference.

\sysname maintains task performance close to or above
\textit{Full history}, with relative gains of 2.9\%--8.5\% in
seven of nine settings, while reducing mean inference cost
by 10.2\%--32.2\% across all nine settings.

\sysname outperforms all five context management baselines in task
performance across all nine settings, with relative gains of
3.3\%--32.5\% in AppWorld TGC and 1.7\%--14.4\% in QA F1.
Mean inference cost is lower in most baseline comparisons.
On AppWorld Test-Challenge with DeepSeek V4 Pro, \sysname incurs
11.7\% higher cost than Semantic Retrieval while achieving
13.6\% higher TGC.

These results show that repeatedly applying \sysname to select
useful tool-result history within a 6K history budget, well below
the models' maximum context windows, achieves performance close
to or above \textit{Full history} at lower inference cost.
Also, compared with other context management baselines, \sysname
consistently improves task performance, with lower cost in
most comparisons.
% Appendix~\ref{app:performance-cost-values} reports the numerical
% performance and cost values.

\begin{table*}[t]
\centering
\scriptsize
\setlength{\tabcolsep}{2.5pt}
\renewcommand{\arraystretch}{1.12}
\caption{Relative changes in task performance and mean inference
cost of \sysname versus each baseline and \textit{Full history}.
Green, red, and gray arrows indicate improvements, regressions,
and no change, respectively.}
\label{tab:relative-performance-cost}
\definecolor{crgain}{RGB}{35,132,52}
\definecolor{crloss}{RGB}{195,48,48}
\definecolor{crneutral}{RGB}{130,130,130}
\newcommand{\crbetter}[2]{\textcolor{crgain}{\ensuremath{#1}#2\%}}
\newcommand{\crworse}[2]{\textcolor{crloss}{\ensuremath{#1}#2\%}}
\newcommand{\crsame}[1]{\textcolor{crneutral}{\ensuremath{\rightarrow}#1\%}}
\resizebox{\textwidth}{!}{%
\begin{tabular}{@{}ll|rr|rr|rr|rr|rr|rr@{}}
\toprule
\textbf{Model} & \textbf{Benchmark}
& \multicolumn{2}{c|}{\textbf{Full history}}
& \multicolumn{2}{c|}{\textbf{Prune}}
& \multicolumn{2}{c|}{\textbf{Compact}}
& \multicolumn{2}{c|}{\shortstack{\textbf{Recency}\\\textbf{+ Summary}}}
& \multicolumn{2}{c|}{\shortstack{\textbf{Semantic}\\\textbf{Retrieval}}}
& \multicolumn{2}{c}{\textbf{ACON}} \\
\cmidrule(lr){3-4}\cmidrule(lr){5-6}\cmidrule(lr){7-8}
\cmidrule(lr){9-10}\cmidrule(lr){11-12}\cmidrule(lr){13-14}
& & \textbf{Perf.} & \textbf{Cost}
& \textbf{Perf.} & \textbf{Cost}
& \textbf{Perf.} & \textbf{Cost}
& \textbf{Perf.} & \textbf{Cost}
& \textbf{Perf.} & \textbf{Cost}
& \textbf{Perf.} & \textbf{Cost} \\
\midrule
\multirow{3}{*}{GPT-4.1} & AppWorld Normal
& \crworse{\downarrow}{2.3} & \crbetter{\downarrow}{32.2} & \crbetter{\uparrow}{14.4} & \crbetter{\downarrow}{3.2} & \crbetter{\uparrow}{12.4} & \crbetter{\downarrow}{0.8} & \crbetter{\uparrow}{7.6} & \crbetter{\downarrow}{1.6} & \crbetter{\uparrow}{7.6} & \crbetter{\downarrow}{18.4} & \crbetter{\uparrow}{3.3} & \crbetter{\downarrow}{14.9} \\

 & AppWorld Challenge
& \crbetter{\uparrow}{8.5} & \crbetter{\downarrow}{28.8} & \crbetter{\uparrow}{19.7} & \crbetter{\downarrow}{2.7} & \crbetter{\uparrow}{12.1} & \crbetter{\downarrow}{1.8} & \crbetter{\uparrow}{11.6} & \crbetter{\downarrow}{16.7} & \crbetter{\uparrow}{14.4} & \crbetter{\downarrow}{9.7} & \crbetter{\uparrow}{11.1} & \crbetter{\downarrow}{8.1} \\

 & 8-objective QA
& \crbetter{\uparrow}{4.5} & \crbetter{\downarrow}{10.2} & \crbetter{\uparrow}{3.7} & \crbetter{\downarrow}{2.0} & \crbetter{\uparrow}{2.1} & \crbetter{\downarrow}{4.5} & \crbetter{\uparrow}{9.5} & \crsame{0.0} & \crbetter{\uparrow}{14.4} & \crsame{0.0} & \crbetter{\uparrow}{10.4} & \crbetter{\downarrow}{5.7} \\

\midrule

\multirow{3}{*}{Muse Glimmer 30B} & AppWorld Normal
& \crbetter{\uparrow}{5.8} & \crbetter{\downarrow}{12.4} & \crbetter{\uparrow}{14.2} & \crworse{\uparrow}{1.7} & \crbetter{\uparrow}{7.4} & \crworse{\uparrow}{0.8} & \crbetter{\uparrow}{8.2} & \crbetter{\downarrow}{3.2} & \crbetter{\uparrow}{10.7} & \crbetter{\downarrow}{11.1} & \crbetter{\uparrow}{9.8} & \crbetter{\downarrow}{2.4} \\

 & AppWorld Challenge
& \crbetter{\uparrow}{2.9} & \crbetter{\downarrow}{19.9} & \crbetter{\uparrow}{16.8} & \crbetter{\downarrow}{7.2} & \crbetter{\uparrow}{9.2} & \crbetter{\downarrow}{16.8} & \crbetter{\uparrow}{10.1} & \crbetter{\downarrow}{2.3} & \crbetter{\uparrow}{13.6} & \crbetter{\downarrow}{8.5} & \crbetter{\uparrow}{8.2} & \crbetter{\downarrow}{12.2} \\

 & 8-objective QA
& \crbetter{\uparrow}{4.6} & \crbetter{\downarrow}{17.1} & \crbetter{\uparrow}{6.9} & \crbetter{\downarrow}{8.2} & \crbetter{\uparrow}{4.6} & \crbetter{\downarrow}{13.9} & \crbetter{\uparrow}{9.3} & \crbetter{\downarrow}{9.0} & \crbetter{\uparrow}{3.1} & \crbetter{\downarrow}{5.1} & \crbetter{\uparrow}{1.7} & \crbetter{\downarrow}{11.5} \\

\midrule

\multirow{3}{*}{DeepSeek V4 Pro} & AppWorld Normal
& \crbetter{\uparrow}{3.6} & \crbetter{\downarrow}{18.9} & \crbetter{\uparrow}{5.9} & \crbetter{\downarrow}{2.5} & \crbetter{\uparrow}{4.3} & \crbetter{\downarrow}{8.3} & \crbetter{\uparrow}{4.3} & \crbetter{\downarrow}{1.3} & \crbetter{\uparrow}{9.9} & \crbetter{\downarrow}{3.8} & \crbetter{\uparrow}{4.3} & \crbetter{\downarrow}{11.5} \\

 & AppWorld Challenge
& \crworse{\downarrow}{0.9} & \crbetter{\downarrow}{23.3} & \crbetter{\uparrow}{32.5} & \crworse{\uparrow}{0.6} & \crbetter{\uparrow}{7.8} & \crbetter{\downarrow}{18.5} & \crbetter{\uparrow}{7.1} & \crworse{\uparrow}{1.1} & \crbetter{\uparrow}{13.6} & \crworse{\uparrow}{11.7} & \crbetter{\uparrow}{12.4} & \crbetter{\downarrow}{14.6} \\

 & 8-objective QA
& \crbetter{\uparrow}{7.1} & \crbetter{\downarrow}{10.7} & \crbetter{\uparrow}{9.6} & \crworse{\uparrow}{0.2} & \crbetter{\uparrow}{4.0} & \crbetter{\downarrow}{4.3} & \crbetter{\uparrow}{3.4} & \crbetter{\downarrow}{2.3} & \crbetter{\uparrow}{9.4} & \crworse{\uparrow}{1.4} & \crbetter{\uparrow}{5.4} & \crbetter{\downarrow}{6.8} \\
\bottomrule
\end{tabular}%
}
\end{table*}

\subsection{Ablation Study}
\label{sec:rq2}

To isolate each selection signal, we compare four \sysname variants:
\textbf{CR--Recency} uses result age,
\textbf{CR--Relevance} uses semantic relevance,
\textbf{CR--Recency + Relevance} combines both, and
\textbf{CR--Full (+Reuse)} adds observed reuse.
We evaluate GPT-4.1 on AppWorld Test-Normal and QA at $B=6$K,
keeping all other components fixed.
To measure Later-Used Result Recall, we first record GPT-4.1
executions with the \textit{Full history} baseline. At each recorded
invocation, we apply each variant's selection rule to the preceding
history to determine which tool results it would retain. We count a
tool result as reused when a later assistant turn exactly repeats
one or more qualifying identifiers first introduced by that result.
Appendix~\ref{app:recall-protocol} details the matching rules.

Table~\ref{tab:rq2-overall} shows that adding observed reuse to
CR--Recency + Relevance improves both task performance and
Later-Used Result Recall. Relative gains are 4.1\% in AppWorld
TGC and 4.5\% in QA F1, alongside recall gains of 41.6\%
and 19.6\%, respectively. CR--Full achieves the highest performance
and recall on both benchmarks, although combining recency and
relevance alone does not consistently improve over either signal.
The recall benefit is particularly strong for tool results needed
many turns after they were produced. On AppWorld, CR--Full achieves
a recall rate approximately $3\times$ that of CR--Recency + Relevance
for tool results needed more than 25 turns after they were produced.
Together, these results support observed reuse as an additional
selection signal that helps retain older results needed later and
improves task performance under the same history budget.

\begin{table*}[t]
\caption{Ablation of \sysname with GPT-4.1 on AppWorld Test-Normal
and 8-objective QA at $B=6$K. Recall denotes Later-Used Result Recall.
Result age is the number of turns between a tool result's creation
and its subsequent reuse. All values are percentages.}
\label{tab:rq2-overall}
\centering
\scriptsize
\setlength{\tabcolsep}{1.6pt}
\renewcommand{\arraystretch}{1.08}

\resizebox{\textwidth}{!}{%
\begin{tabular}{@{}lcc@{\hspace{10pt}}cc*{4}{>{\centering\arraybackslash}p{0.10\textwidth}}@{}}
\toprule
\multirow{2}{*}{\textbf{Variant}}
& \multicolumn{2}{c@{\hspace{10pt}}}{\textbf{Performance}}
& \multicolumn{2}{c}{\textbf{Later-Used Result Recall}}
& \multicolumn{4}{c}{\shortstack{
    \textbf{AppWorld Later-Used Result Recall}\\
    \textbf{by Result Age}}} \\
\cmidrule(lr){2-3}
\cmidrule(lr){4-5}
\cmidrule(lr){6-9}
& \shortstack{\textbf{AppWorld}\\\textbf{(TGC)}}
& \shortstack{\textbf{QA}\\\textbf{(F1)}}
& \textbf{AppWorld}
& \textbf{QA}
& $\mathbf{\leq 3}$
& \textbf{4--10}
& \textbf{11--25}
& $\mathbf{>25}$ \\
\midrule

CR--Recency
& 71.4 & 48.5
& 63.8 & 86.0
& 95.7 & 78.7 & 37.7 & 18.5 \\

CR--Relevance
& 73.2 & 47.5
& 71.4 & 64.2
& 89.5 & 81.9 & 54.1 & 47.1 \\

CR--Recency + Relevance
& 72.6 & 50.8
& 61.6 & 76.1
& 84.4 & 72.5 & 43.9 & 22.7 \\

\rowcolor{crblue}
\textbf{CR--Full (+Reuse)}
& \textbf{75.6} & \textbf{53.1}
& \textbf{87.2} & \textbf{91.0}
& \textbf{99.3} & \textbf{91.5}
& \textbf{79.0} & \textbf{68.5} \\

\bottomrule
\end{tabular}%
}
\end{table*}

\subsection{Sensitivity to the History Budget}
\label{sec:budget-sensitivity}

\begin{figure*}[t]
    \centering
    \begin{minipage}[t]{0.49\textwidth}
        \centering
        \includegraphics[width=\linewidth]{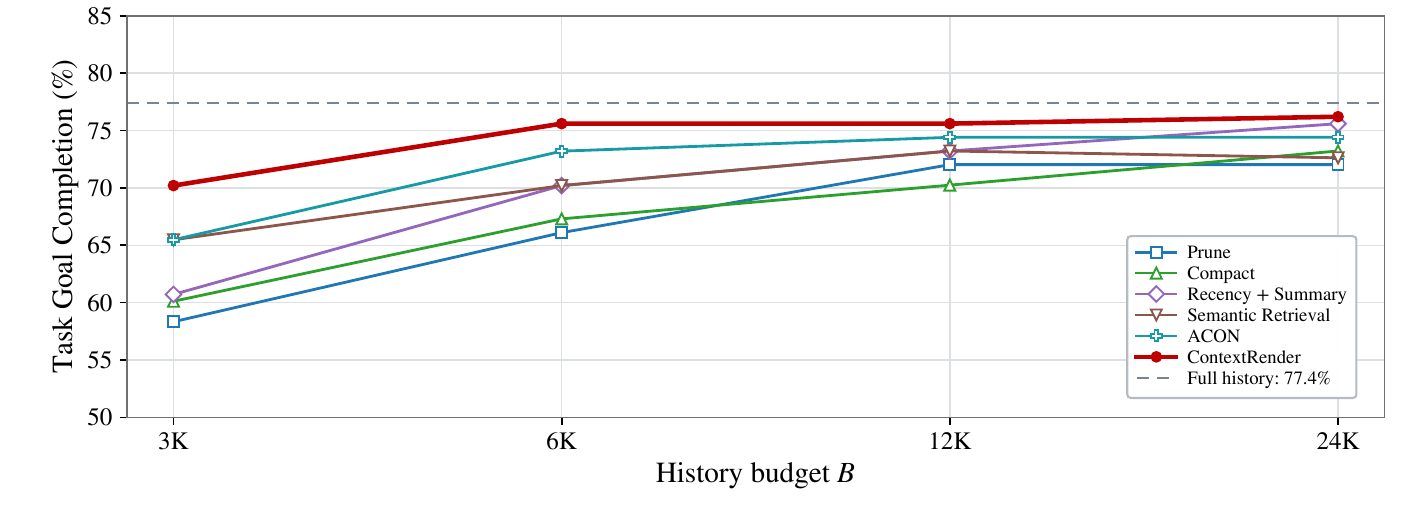}
        \par\smallskip
        {\small (a) AppWorld Test-Normal, GPT-4.1}
    \end{minipage}\hfill
    \begin{minipage}[t]{0.49\textwidth}
        \centering
        \includegraphics[width=\linewidth]{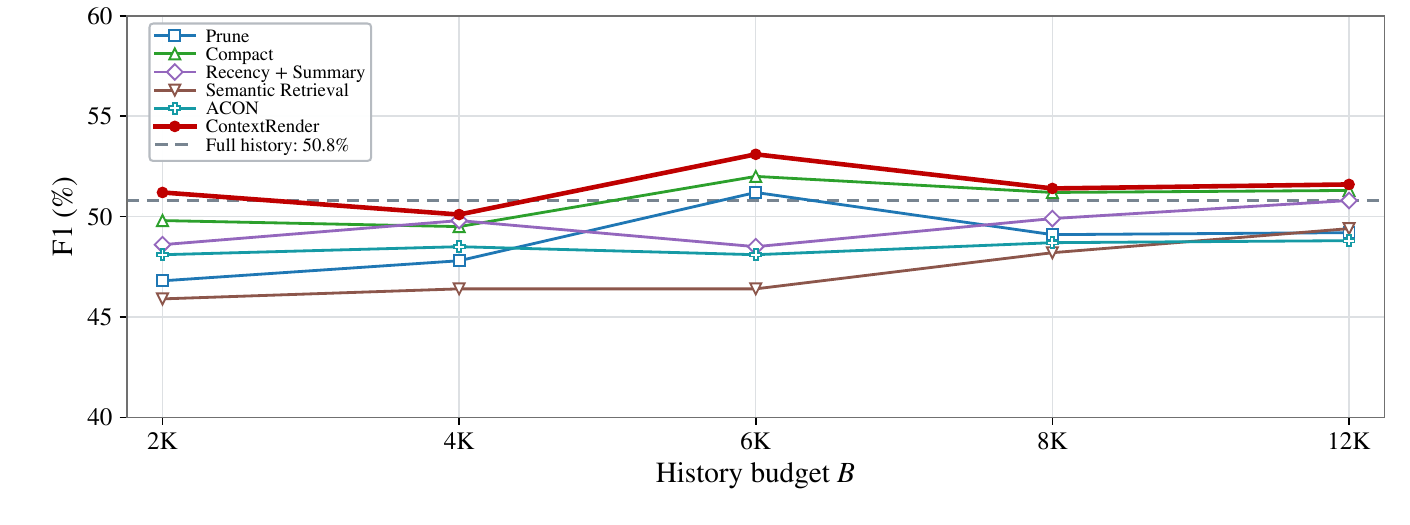}
        \par\smallskip
        {\small (b) 8-objective QA, GPT-4.1}
    \end{minipage}
    \caption{Budget sensitivity with GPT-4.1. $B$ denotes the history budget. ``Recency'' denotes Recency + Summary.}
    \label{fig:rq1-budget-sweep}
    \vspace{-0.2cm}
\end{figure*}

Figure~\ref{fig:rq1-budget-sweep} varies the history budget from
3K to 24K on AppWorld Test-Normal and from 2K to 12K on QA.
\sysname outperforms the strongest evaluated context management
baseline at every tested budget, with gains of up to 4.7 percentage
points in AppWorld TGC and 1.4 percentage points in QA F1.
On AppWorld, performance moves closer to \textit{Full history}
as the budget increases, with little further improvement beyond
6K. QA performance varies within a 3.0-point range, remaining
close to or above \textit{Full history}. These results show that
\sysname maintains its advantage across budget settings and
achieves strong task performance with a small history budget.

\section{Related Work}
\label{sec:related}

% \paragraph{Context compression and masking.}
ACON optimizes compression guidelines; Context-Folding and
AgentFold condense interaction histories
\citep{kang2025acon,sun2025contextfolding,ye2025agentfold}.
MEM1 learns a compact internal state, while observation masking
hides earlier outputs without summaries
\citep{zhou2025mem1,lindenbauer2025complexitytrap,zhang2026masking}.
These reductions can omit details needed later, which cannot
be recovered from the reduced history alone if originals are
discarded. \sysname retains complete tool results and selects
which to expose at each invocation, allowing omitted information
to return.

% \paragraph{Adaptive and recoverable context.}
Adaptive Context Elasticizer uses model decisions to expose,
abstract, or hide historical interactions, while PACE adjusts
their granularity using semantic relevance
\citep{liao2026ace,wei2026pace}. Sculptor provides context
editing and restoration tools, and Addressable Recall Compaction
supports recall by identifier
\citep{li2025sculptor,dang2026arc}.
\sysname adds observed reuse as selection evidence. Tool-Flow
Analysis attributes later uses of distinctive references to
originating tool results. The renderer combines this evidence
with recency and semantic relevance to select results within
the historical budget.

% \paragraph{Agent memory and context adaptation.}
MemGPT manages active context and external memory; A-MEM links
memory notes through semantic retrieval and model analysis
\citep{packer2023memgpt,xu2025amem}. Agentic Context Engineering
develops reusable playbooks through generation, reflection,
and curation \citep{zhang2025ace}.
\sysname selects tool results accumulated within the current
task. Its graph records chronological order and observed reuse
by later operations, providing evidence of continued usefulness
beyond content similarity. This complements memory organization
and the adaptation of reusable guidance.

\section{Conclusion}

We introduced \sysname{}, which combines observed reuse, recency, and semantic relevance to select historical tool results from a persistent graph while retaining omitted results for later use. Across AppWorld and 8-objective QA with three execution models, it outperforms evaluated managed-context baselines using a 6K history budget, well below the models' maximum context windows. It maintains task performance close to or above \textit{Full history} while reducing mean inference cost by 10.2\%--32.2\%.

\subsection*{AI use statement}
We used generative AI tools to assist with polishing text and formatting experimental tables. All experimental design, measurements, and analyses were conducted by the authors. We reviewed all AI-assisted content and take full responsibility for the final content of this work.

% \subsection*{Reproducibility statement}
% We provide implementation details, model and baseline configurations,
% context-selection parameters, evaluation protocols and runtime integration
% details in the appendix.
% These details are intended to support reproduction of the main
% experiments and ablations.
% TODO(before ICLR submission): point to the released server, adapters, suites,
% pre-registration documents, and analysis scripts.

% \bibliography{references}
% \bibliographystyle{iclr2027_conference}

\bibliography{references}

@article{kang2025acon,
  title={ACON: Optimizing Context Compression for Long-horizon LLM Agents},
  author={Kang, Minki and Chen, Wei-Ning and Han, Dongge and Inan, Huseyin A. and Wutschitz, Lukas and Chen, Yanzhi and Sim, Robert and Rajmohan, Saravan},
  journal={arXiv preprint arXiv:2510.00615},
  year={2025}
}

@article{lindenbauer2025complexitytrap,
  title={The Complexity Trap: Simple Observation Masking Is as Efficient as {LLM} Summarization for Agent Context Management},
  author={Lindenbauer, Tobias and Slinko, Igor and Felder, Ludwig and Bogomolov, Egor and Zharov, Yaroslav},
  journal={arXiv preprint arXiv:2508.21433},
  year={2025}
}

@article{packer2023memgpt,
  title={{MemGPT}: Towards {LLM}s as Operating Systems},
  author={Packer, Charles and Wooders, Sarah and Lin, Kevin and Fang, Vivian and Patil, Shishir G. and Stoica, Ion and Gonzalez, Joseph E.},
  journal={arXiv preprint arXiv:2310.08560},
  year={2023}
}

@article{xu2025amem,
  title={{A-MEM}: Agentic Memory for {LLM} Agents},
  author={Xu, Wujiang and Liang, Zujie and Mei, Kai and Gao, Hang and Tan, Juntao and Zhang, Yongfeng},
  journal={arXiv preprint arXiv:2502.12110},
  year={2025},
  url={https://arxiv.org/abs/2502.12110}
}

@article{zhang2025ace,
  title={Agentic Context Engineering: Evolving Contexts for Self-Improving Language Models},
  author={Zhang, Qizheng and Hu, Changran and Upasani, Shubhangi and Ma, Boyuan and Hong, Fenglu and Kamanuru, Vamsidhar and Rainton, Jay and Wu, Chen and Ji, Mengmeng and Li, Hanchen and Thakker, Urmish and Zou, James and Olukotun, Kunle},
  journal={arXiv preprint arXiv:2510.04618},
  year={2025},
  url={https://arxiv.org/abs/2510.04618}
}

@inproceedings{trivedi2024appworld,
  title={{AppWorld}: A Controllable World of Apps and People for Benchmarking Interactive Coding Agents},
  author={Trivedi, Harsh and Khot, Tushar and Hartmann, Mareike and Manku, Ruskin and Dong, Vinty and Li, Edward and Gupta, Shashank and Sabharwal, Ashish and Balasubramanian, Niranjan},
  booktitle={Proceedings of the 62nd Annual Meeting of the Association for Computational Linguistics (Volume 1: Long Papers)},
  year={2024}
}

@article{liu2024lostinthemiddle,
  title={Lost in the Middle: How Language Models Use Long Contexts},
  author={Liu, Nelson F. and Lin, Kevin and Hewitt, John and Paranjape, Ashwin and Bevilacqua, Michele and Petroni, Fabio and Liang, Percy},
  journal={Transactions of the Association for Computational Linguistics},
  volume={12},
  pages={157--173},
  year={2024}
}

@article{yang2024sweagent,
  title={{SWE}-agent: Agent-Computer Interfaces Enable Automated Software Engineering},
  author={Yang, John and Jimenez, Carlos E. and Wettig, Alexander and Lieret, Kilian and Yao, Shunyu and Narasimhan, Karthik and Press, Ofir},
  journal={arXiv preprint arXiv:2405.15793},
  year={2024}
}

@article{ye2025agentfold,
  title={{AgentFold}: Long-Horizon Web Agents with Proactive Context Management},
  author={Ye, Rui and Zhang, Zhongwang and Li, Kuan and Yin, Huifeng and Tao, Zhengwei and Zhao, Yida and Su, Liangcai and Zhang, Liwen and Qiao, Zile and Wang, Xinyu and Xie, Pengjun and Huang, Fei and Chen, Siheng and Zhou, Jingren and Jiang, Yong},
  journal={arXiv preprint arXiv:2510.24699},
  year={2025}
}

@article{sun2025contextfolding,
  title={Scaling Long-Horizon {LLM} Agent via Context-Folding},
  author={Sun, Weiwei and Lu, Miao and Ling, Zhan and Liu, Kang and Yao, Xuesong and Yang, Yiming and Chen, Jiecao},
  journal={arXiv preprint arXiv:2510.11967},
  year={2025}
}

@article{dang2026arc,
  title={Addressable Recall Compaction for Long Context-Window Control in {AI} Agents},
  author={Dang, Thang and Ichikawa, Yuma and Fatima, Sakina and Shirahata, Koichi},
  journal={arXiv preprint arXiv:2607.25066},
  year={2026},
  doi={10.48550/arXiv.2607.25066},
  url={https://arxiv.org/abs/2607.25066}
}

@article{li2025sculptor,
  title={Sculptor: Empowering {LLMs} with Cognitive Agency via Active Context Management},
  author={Li, Mo and Xu, L.H. and Tan, Qitai and Cao, Ting and Liu, Yunxin},
  journal={arXiv preprint arXiv:2508.04664},
  year={2025}
}

@article{zhang2026masking,
  title={Masking Stale Observations Helps Search Agents -- Until It Doesn't: A Regime Map and Its Mechanism},
  author={Zhang, Haoxiang and Xu, Qixin and Li, Zhuofeng and Zhang, Lei and Jiang, Pengcheng and Zhang, Yu and McAuley, Julian},
  journal={arXiv preprint arXiv:2606.00408},
  year={2026}
}

@article{liao2026ace,
  title={{ACE}: Pluggable Adaptive Context Elasticizer across Agents},
  author={Liao, Ning and Long, Zihao and Wang, Xiaoxing and Yang, Xue and Wang, Yaoming and Zhuang, Ziyuan and Cai, Xunliang and Weng, Rongxiang and Yan, Junchi},
  journal={arXiv preprint arXiv:2606.31564},
  year={2026}
}

@inproceedings{wei2026pace,
  title={{PACE}: Predictive Adaptive Context Extraction for Long-Horizon {LLM} Agents},
  author={Wei, Lei and Peng, Xiao and Tt and Zhang, Guannan and Jiang, Chenhao and Li, Hongyu and Lin, Lanbo and Xu, Yuanwu and Liu, Jiayao and Wang, Kesu and Wang, Bin},
  booktitle={Proceedings of the 64th Annual Meeting of the Association for Computational Linguistics (Volume 1: Long Papers)},
  pages={27184--27199},
  publisher={Association for Computational Linguistics},
  address={San Diego, California, United States},
  year={2026},
  doi={10.18653/v1/2026.acl-long.1252},
  url={https://aclanthology.org/2026.acl-long.1252/}
}

@article{hao2026selfgc,
  title={{Self-GC}: Self-Governing Context for Long-Horizon {LLM} Agents},
  author={Hao, Xubin and Meng, Hongjin and Yin, Xin and Zhu, Jiawei and Cao, Chenpeng},
  journal={arXiv preprint arXiv:2607.00692},
  year={2026}
}

@article{kwiatkowski2019nq,
  title={Natural Questions: A Benchmark for Question Answering Research},
  author={Kwiatkowski, Tom and Palomaki, Jennimaria and Redfield, Olivia and Collins, Michael and Parikh, Ankur and Alberti, Chris and Epstein, Danielle and Polosukhin, Illia and Devlin, Jacob and Lee, Kenton and Toutanova, Kristina and Jones, Llion and Kelcey, Matthew and Chang, Ming-Wei and Dai, Andrew M. and Uszkoreit, Jakob and Le, Quoc and Petrov, Slav},
  journal={Transactions of the Association for Computational Linguistics},
  volume={7},
  pages={452--466},
  year={2019}
}

@article{zhou2025mem1,
  title={{MEM1}: Learning to Synergize Memory and Reasoning for Efficient Long-Horizon Agents},
  author={Zhou, Zijian and Qu, Ao and Wu, Zhaoxuan and Kim, Sunghwan and Prakash, Alok and Rus, Daniela and Zhao, Jinhua and Low, Bryan Kian Hsiang and Liang, Paul Pu},
  journal={arXiv preprint arXiv:2506.15841},
  year={2025},
  url={https://arxiv.org/abs/2506.15841}
}

@misc{smith2025openhands,
  title={{OpenHands} Context Condensation for More Efficient {AI} Agents},
  author={Smith, Calvin},
  howpublished={All Hands AI Blog, \url{https://www.all-hands.dev/blog/openhands-context-condensensation-for-more-efficient-ai-agents}},
  month=apr,
  year={2025}
}

@article{mars2023jaseci,
  author={Jason Mars and Yiping Kang and Roland Daynauth and Baichuan Li
          and Ashish Mahendra and Krisztian Flautner and Lingjia Tang},
  title={The Jaseci Programming Paradigm and Runtime Stack:
         Building Scale-Out Production Applications Easy and Fast},
  journal={IEEE Computer Architecture Letters},
  volume={22},
  number={2},
  pages={101--104},
  year={2023},
  doi={10.1109/LCA.2023.3274038}
}

@article{mars2025osp,
  author={Jason Mars},
  title={Object-Spatial Programming},
  journal={arXiv preprint arXiv:2503.15812},
  year={2025}
}

@misc{openclaw2026,
  author       = {{OpenClaw}},
  title        = {{OpenClaw}},
  year         = {2026},
  howpublished = {Open-source software},
  url          = {https://github.com/openclaw/openclaw},
  note         = {Accessed September 25, 2026}
}

@misc{opencode2025,
  author       = {{OpenCode}},
  title        = {{OpenCode}},
  year         = {2025},
  howpublished = {Open-source software},
  url          = {https://github.com/anomalyco/opencode},
  note         = {Accessed September 25, 2026}
}

@misc{hermes2026,
  author       = {{Nous Research}},
  title        = {{Hermes Agent}},
  year         = {2026},
  howpublished = {Open-source software},
  url          = {https://github.com/NousResearch/hermes-agent},
  note         = {Accessed September 25, 2026}
}
\bibliographystyle{iclr2027_conference}

\clearpage
\appendix
\appendix

\section{Implementation Details}
\label{app:implementation}

This appendix describes the execution procedure, TFA matching rules,
scoring configuration, and history-budget accounting. Evaluation
settings and protocols appear in Appendix~\ref{app:additional-evaluation},
cost accounting and numerical results in Appendix~\ref{app:cost-results},
and runtime integrations in Appendix~\ref{app:adapters}.

\subsection{Execution Procedure}
\label{app:procedure}

Algorithm~\ref{alg:context-rendering} summarizes the rendering
procedure in Section~\ref{sec:rendering}. The server updates the
persistent graph before each invocation. If the accumulated
execution history fits within $B$, it is included in full.
Otherwise, the renderer scores historical tool results and greedily
selects those that fit. The server returns the rendered history to
the runtime adapter.

\begin{algorithm}[t]
\caption{Context-management pass before invocation $t$}
\label{alg:context-rendering}
\begin{algorithmic}[1]
\Require Persistent graph $G$, new interaction events $E_t$,
         history budget $B$
\Ensure Rendered execution history $H_t$ with $\operatorname{Tok}(H_t)\leq B$
\State Add previously unseen events from $E_t$ to $G$
\State Detect new reuse matches and add their attributed reuse edges to $G$
\State Let $H_t^{\mathrm{full}}$ be the accumulated execution history
\If{$\operatorname{Tok}(H_t^{\mathrm{full}})\leq B$}
    \State \Return $H_t^{\mathrm{full}}$
\EndIf
\State Construct the relevance vector $q_t$
\For{each historical tool result $r$ in $G$}
    \State Compute $c_t(r)$ using Equation~\ref{eq:reuse-mass}
    \State Compute $U_t(r)$ using Equation~\ref{eq:usefulness}
\EndFor
\State Apply the oversized-result fallback where needed (Appendix~\ref{app:budget-accounting})
\State $S_t\gets\varnothing$
\While{an unselected result satisfies Equation~\ref{eq:history-feasibility}}
    \State Choose the feasible result with maximum $R_t(r\mid S_t)$
    \State Break exact ties by transcript order, older first
    \State Add the chosen result to $S_t$
\EndWhile
\State $H_t\gets\mathcal{H}_t(S_t)$ in chronological order
\State \Return $H_t$ to the runtime adapter
\end{algorithmic}
\end{algorithm}

Unselected results remain in the persistent graph with their
complete content and can be reconsidered at subsequent invocations.

\subsection{Tool-Flow Analysis Implementation}
\label{app:reference-matching}

\paragraph{Trace model and value reuse.}
TFA analyzes the realized execution trace as a straight-line sequence
of events. The extraction and filtering rules below define its
\emph{value abstraction}. Under this abstraction, a tool result
defines the set of qualifying values it contains. A later model
output or tool call uses a value when it contains an exact,
case-sensitive whole-token match to that value.

For each used value, TFA records a \emph{reuse link} connecting the
use to the first earlier tool result containing that value, as in
Section~\ref{sec:usefulness}. Intermediate copies in model outputs
or tool calls do not change the attributed source. These links
form the reuse edges in the persistent context graph. Because the
analysis follows a single realized trace, dependency tracking
reduces to constructing these reuse links; no merging across
alternative control-flow paths is required.

\paragraph{Value abstraction and matching rules.}
TFA extracts candidate values using the pattern
\verb|[A-Za-z0-9][A-Za-z0-9_.\-]{4,}|. Quote marks, bracket
characters, commas, colons, slashes, and whitespace act as separators.
Leading and trailing periods, hyphens, and underscores are removed
from each candidate, while the same characters are retained when they
occur inside the value. Candidates shorter than six characters after
this stripping step are discarded.

A candidate is retained only if it contains at least one underscore,
uppercase letter, digit, hyphen, or period. Pure-alphabetic Titlecase
words are excluded. Matching is case-sensitive and requires an exact
whole-token match; no case normalization or substring matching is
performed.

File paths are not parsed separately. Because slashes act as
separators, each path segment is tested independently using the same
extraction and filtering rules. Retained values are matched against
later model outputs and tool calls.

\paragraph{Relation to the motivating example.}
Figure~2 illustrates task-level information dependencies, not a
literal set of TFA-detected edges. The displayed names do not qualify:
\texttt{Kathryn} is excluded as Titlecase, while \texttt{Troy} and
\texttt{Nancy} are shorter than six characters. Email addresses are
split at \texttt{@}; \texttt{kathrynmaldonado} fails the character
filter, whereas \texttt{tr\_solo}, \texttt{nan\_ritt}, and
\texttt{gmail.com} qualify.

Nancy's email is not reused before Turn~18. The match to Troy's
\texttt{tr\_solo} at Turn~16 instead credits the shared Turn~8 tool
result. This can raise the priority of the whole contact result,
helping retain Nancy's information for later use without an earlier
match to her email.

\paragraph{Scope of the reuse signal.}
We use a result's past reuse as a heuristic predictor of future
usefulness. A match to one value increases the score of its source
result and can therefore help retain other values in that result.
This relies on information locality within a tool result; it does
not establish a dependency on the particular value needed by an
upcoming operation. TFA records observed value reuse and attributes
it to earlier tool results; the renderer uses these reuse edges to
prioritize complete results.

\paragraph{Attribution and provenance.}
For production TFA, the source of a matched value is the first earlier
tool result containing it. References to a result that ``introduces''
a value or to an ``originating'' result use this tool-result attribution
rule: ``first'' is measured across earlier tool results.
Under this rule, a qualifying value
already present in the task input or an earlier assistant message can
be attributed to the first tool result that echoes it when the value
is matched in a later operation. This attribution supplies an online
ranking signal; it does not establish either the original source of a
value or a causal dependency.
The recall protocol in Appendix~\ref{app:recall-protocol} requires a
token's first transcript appearance to occur in a tool result to
label events for result-availability evaluation.

\paragraph{Match counting and weighting.}
Each distinct qualifying value counts once per later model output
or tool call. Repeated occurrences of the same value within one
operation do not create additional matches. Uses in separate
operations count separately, and different qualifying values within
one operation contribute separate matches.

Before invocation $t$, $\mathcal{M}_t(r)$ contains the matches
attributed to result $r$. The count $n_t(m)$ is the number of
historical tool results containing the value in match $m$ at that
invocation, including results outside the current model input.
Equation~\ref{eq:reuse-mass} weights each match by $1/n_t(m)$.
This count is over tool results, not occurrences within a result:
three contacts sharing \texttt{gmail.com} in one result contribute
one to $n_t(m)$. Occurrences in other results increase the count.
The weighting reduces a match's score contribution without removing
the match. The renderer recomputes $c_t(r)$ using only the graph
state available before invocation $t$.

% \paragraph{Scope of the signal.}
% These rules record matching values as evidence for context selection.
% They can miss values whose form changes, and a value shared by
% several results does not establish which result supplied a later use.
% First-source attribution is an operational rule for assigning credit.
% The signal does not recover every execution dependency.
% The rules used to measure Later-Used Result Recall in
% Appendix~\ref{app:recall-protocol} are separate from these production
% TFA rules.

\subsection{Usefulness Configuration}
\label{app:usefulness-configuration}

The usefulness score combines result age, semantic relevance, and
observed reuse. Following Section~\ref{sec:rendering}, one turn
corresponds to one model invocation, and $a_t(r)$ counts the turns
since result $r$ was produced. Thus, $\lambda=0.3$ controls recency
decay per model invocation. Assistant and tool messages within a
turn do not separately increment this age.

For semantic relevance, let $x$ denote the task's first user message
and let $f_t$ denote the latest visible interaction before invocation
$t$, formed by concatenating the latest visible user message and latest
visible assistant message. We construct
\begin{equation}
q_t
=
0.4\,\operatorname{unit}(\operatorname{emb}(x))
+
0.6\,\operatorname{unit}(\operatorname{emb}(f_t)),
\label{eq:relevance-query}
\end{equation}
where $\operatorname{unit}(\cdot)$ denotes vector normalization.
The first user message is clipped to 2,000 characters before embedding.

We use \texttt{text-embedding-3-small} through OpenAI via LiteLLM,
embedding the first 8,000 characters of each tool result.

Semantic relevance is
\begin{equation}
s_t(r)
=
\max\!\left(
\cos(\operatorname{emb}(r),q_t),0
\right).
\end{equation}

\paragraph{Parameters.}
We use
\begin{align*}
W_{\mathrm{rec}}   &= 1.0, &
W_{\mathrm{rel}}   &= 1.0, &
W_{\mathrm{reuse}} &= 1.0, \\
\lambda            &= 0.3, &
\gamma             &= 0.2.
\end{align*}
The parameter $\lambda$ controls recency decay with result age, while
$\gamma$ controls how quickly the reuse contribution increases as
observed reuse accumulates. Equal coefficients do not imply equal
per-result contributions because the three signals can take different
values.

\subsection{Selection and Budget Accounting}
\label{app:budget-accounting}

\paragraph{History budget.}
The history budget $B$ strictly caps the rendered execution history,
as described in Section~\ref{sec:rendering}. If the accumulated
history fits within $B$, it is included in full. Otherwise, the
renderer selects historical tool results. Accounting includes
retained history messages, associated tool calls, and selected
result text. The system prompt and current user message are separate
model-input components, so $B$ is distinct from the model's total
context window. We count history tokens using \texttt{tiktoken}.

\paragraph{Other historical messages.}
Historical user and assistant messages are retained verbatim in
chronological order, excluding the current user message accounted
for separately above. Their tokens are charged to $B$ before
allocating space to selected tool results and their associated
tool calls. A tool call already contained in a retained assistant
message is counted once, as part of the assembled history.

\paragraph{Budget feasibility.}
Let $S$ be the selected results and $\mathcal{H}_t(S)$ the history
assembled from their text, associated tool calls, and retained
history messages. A candidate $r$ is feasible only if
\begin{equation}
\operatorname{Tok}\!\left(\mathcal{H}_t(S\cup\{r\})\right)\leq B.
\label{eq:history-feasibility}
\end{equation}
Thus, feasibility is checked against the assembled history, not
only the candidate's result text. Retained non-tool history is
included in every feasibility check, including before any tool
results are selected.

\paragraph{Greedy selection.}
Selection starts with $S=\varnothing$. The renderer ranks feasible
candidates by their usefulness from Equation~\ref{eq:usefulness},
adjusted for similarity to results already selected:
\begin{equation}
\begin{split}
R_t(r\mid S) &= U_t(r)-0.5\,D(r,S),\\
D(r,S) &= \max\!\left(0,\max_{r'\in S}
\cos\!\left(\operatorname{emb}(r),\operatorname{emb}(r')\right)
\right).
\end{split}
\label{eq:selection-penalty}
\end{equation}
Here $D(r,\varnothing)=0$. We set the coefficient to $0.5$.
The floor at zero prevents negative cosine similarity from
increasing a candidate's score. The embeddings are specified in
Appendix~\ref{app:usefulness-configuration}.

At each step, the renderer adds the feasible result with the
highest $R_t(r\mid S)$. Candidates are visited in transcript order,
and the current best is replaced only by a strictly higher score;
exact ties therefore favor the older result. A result that does not
fit is skipped while other candidates remain eligible. There is no
minimum score, and selection ends when no remaining result fits.

\paragraph{Oversized-result fallback.}
If an individual tool result exceeds $B$, \texttt{gpt-4.1} compacts
it. The compacted text is included only if the assembled history still
satisfies Equation~\ref{eq:history-feasibility}; its complete
original remains stored in the graph. We did not observe such
oversized individual tool results in the evaluation runs, where
selected results were included in full.

\paragraph{Rendered output.}
Selected results and their associated tool calls are returned in
chronological order, rather than score order. Unselected results
are omitted from the model-visible history and remain stored in
full in the persistent graph. Before subsequent invocations, their
scores are recomputed from the updated graph, allowing them to be
selected again as priorities change. The rendered history always
satisfies $\operatorname{Tok}(H_t)\leq B$.

\section{Evaluation Settings and Protocols}
\label{app:additional-evaluation}

\subsection{Model and API Settings}
\label{app:model-settings}

We evaluate \texttt{gpt-4.1} through the OpenAI API,
\texttt{muse-glimmer:30b-q8\_0} using local Ollama 0.34.0,
and \texttt{deepseek-v4-pro:0813} through Ollama Cloud.
We set temperature to $0.0$, top-$p$ to $1.0$, and seed to
$42$. Each method is executed once per task.

The primary setting uses $B=6$K. The GPT-4.1 budget-sensitivity
study uses $B\in\{3,6,12,24\}$K on AppWorld Test-Normal and
$B\in\{2,4,6,8,12\}$K on 8-objective QA. For each benchmark
and execution model, methods use the same agent prompt, tool
interface, and execution limits.

\subsection{Baseline Implementation Details}
\label{app:baseline-details}

All budgeted context management methods use the same history budget
$B$, tokenizer-based counting, and budget scope within each model
and benchmark. The budget covers retained historical messages,
tool calls, tool results, and summaries, with the same prompt
exclusions described in Appendix~\ref{app:budget-accounting}.
The history cap is distinct from the complete model-input length.

\paragraph{Prune.}
Prune considers complete historical tool results from newest to oldest and
retains those that fit within $B$. Results outside the budget are omitted
from the context presented to the model.

\paragraph{Compact (LLM).}
When the accumulated interaction history exceeds $B$, Compact (LLM)
uses an LLM to compress the entire interaction history to fit within
the budget. It uses
\texttt{gpt-4.1} through LiteLLM, configured by
\texttt{CTXGRAPH\_MODEL}. The summary prompt is generated by the
byLLM typed-slot extractor and organizes information into files read,
key observations, commands run, decisions, and unresolved questions.
It asks the model to preserve numeric values, configuration keys, paths,
identifiers, and error strings verbatim.

\paragraph{Recency + Summary.}
Recency + Summary retains recent complete tool results and summarizes
older results within the same budget $B$. It uses \texttt{gpt-4.1}
through LiteLLM with the same summary prompt as Compact (LLM).
When an earlier summary exists, the summarizer merges it with newly
summarized content.

\paragraph{Semantic Retrieval.}
Semantic Retrieval uses the embedding model configured by
\texttt{CTXGRAPH\_EMBED\_MODEL}, with \texttt{text-embedding-3-small} as the
default, through LiteLLM. Before each model invocation, it forms the
retrieval input from the latest user and assistant messages. Results
from the three most recent turns are considered first, newest first,
using up to 60\% of $B$. Remaining results are ranked by cosine
similarity to this input and included when they fit within the
remaining budget. Unselected results remain stored and can be
selected again at a later invocation.

\paragraph{ACON.}
We use the authors' released history compression implementation with their
optimized compression guideline. The compressor is \texttt{gpt-4.1} on both
AppWorld and 8-objective QA. The history budget is the same
$B$ used by the other context management methods.

\paragraph{Full-history reference.}
We also evaluate \textit{Full history}, which passes the entire
accumulated execution history without context management even when
it exceeds $B$, as a reference for task performance and inference
cost.

\subsection{Scoring Parameters and Ablation Configuration}
\label{app:scoring-ablation}

\paragraph{Ablation configuration.}
We construct selector ablations by setting omitted scoring terms in
Equation~\ref{eq:usefulness} to zero. Retained terms keep their
full-selector weights without renormalization or retuning. The values
of $\lambda$ and $\gamma$ remain fixed wherever their corresponding
terms are active. Table~\ref{tab:ablation-weights} lists the resulting
weights.

\begin{table}[t]
\centering
\small
\setlength{\tabcolsep}{6pt}
\renewcommand{\arraystretch}{1.08}
\caption{Scoring weights used in the selector ablations. Zero disables
the corresponding term; retained weights remain unchanged.}
\label{tab:ablation-weights}
\begin{tabular}{lccc}
\toprule
\textbf{Selector}
& $W_{\mathrm{rec}}$
& $W_{\mathrm{rel}}$
& $W_{\mathrm{reuse}}$ \\
\midrule
CR--Recency             & 1.0 & 0.0 & 0.0 \\
CR--Relevance           & 0.0 & 1.0 & 0.0 \\
CR--Recency + Relevance & 1.0 & 1.0 & 0.0 \\
CR--Full (+Reuse)       & 1.0 & 1.0 & 1.0 \\
\bottomrule
\end{tabular}
\end{table}

All variants use the same persistent graph, history budget,
budget accounting, and greedy selection procedure in
Appendix~\ref{app:budget-accounting}. Each starts with an empty
selected set and applies the same similarity adjustment. They differ
only in the enabled terms of $U_t(r)$: setting $W_{\mathrm{reuse}}=0$
removes observed reuse from selection, with no separate
identifier-based protection. The comparison measures the contribution
of the scoring terms under this shared configuration.
% It does not separately isolate TFA's extraction, attribution, or
% inverse-frequency weighting rules.

\paragraph{Interpreting combined signals.}
Combining signals changes the usefulness scores of results competing
for the available budget. A result favored by one signal can be
displaced when another signal increases the usefulness of competing
results. Consequently, Recency + Relevance need not outperform either
signal alone under fixed weights. The observed ordering characterizes
this configuration and does not establish the ordering that would
result from separately tuned variants.

\subsection{Later-Used Result Recall Protocol}
\label{app:recall-protocol}

\paragraph{Reuse events.}
We define lexical reuse events by exact token matching. Eligible tokens
consist of at least six letters, digits, or underscores and include at least
one uppercase letter or underscore. A token is linked to a tool result
only when its first appearance in the transcript occurs in that result
rather than in the task input or an earlier assistant message. A later
assistant turn containing the token, including in prose, code, or
tool-call arguments, labels a lexical reuse event.

\paragraph{Relation to production matching.}
Production TFA uses value-level matches to rank results during
execution. Recall evaluation uses a fixed lexical event set on
recorded traces to compare result availability across selectors.
The separate labeling rule provides a shared evaluation proxy whose
events do not depend on each selector's production TFA detections.

Unlike production TFA (Appendix~\ref{app:reference-matching}), the
recall rule permits Titlecase words of sufficient length, requires an
uppercase letter or underscore, and limits eligible tokens to letters,
digits, and underscores. Its transcript-level first-appearance rule
focuses evaluation on values first observed in tool results, excluding
values already supplied by the task or earlier assistant text.
Production attribution instead assigns matches to the first tool
result containing the value.

Each distinct pair $(j,r)$ of assistant turn $j$ and earlier tool
result $r$ counts once, regardless of the number of matched tokens.
This counts availability once per result at each turn, so multiple
identifiers from one result do not inflate its contribution.
Production TFA instead counts distinct values per later operation
as evidence for its reuse score. Reuse of the same result at different
turns produces separate evaluation events. The recall age bins below
count turns in model invocations, using the same turn definition as
Section~\ref{sec:rendering} and
Appendix~\ref{app:usefulness-configuration}.

\paragraph{Visibility.}
For each event $(j,r)$, result $r$ is visible only if its complete
original text is present in the model-visible context immediately
before turn $j$. An LLM-compacted version does not count as
complete-result visibility. A previously omitted result counts
if it has been selected again
and restored in full before the evaluated turn. If $E$ is the event
set and $V_j$ is the set of results visible in full before turn $j$,
we compute
\begin{equation}
\mathrm{Recall}
=
100\,
\frac{
\sum_{(j,r)\in E}\mathbf{1}[r\in V_j]
}{
|E|
}.
\label{eq:later-used-result-recall}
\end{equation}

\paragraph{Measurement procedure.}
We first record GPT-4.1 executions using the \textit{Full history}
reference. At each recorded invocation, we apply each variant's
selection rule to the preceding history, as described in
Section~\ref{sec:rq2}.
Before assistant turn $j$, the evaluator synchronizes all messages
preceding $j$, obtains the selector's render decision, and checks
whether result $r$ is visible. Turn $j$ is used only afterward to
label the event. Reuse counters use only turns preceding $j$.
Consequently, every selector is evaluated on the same event set.

\paragraph{Interpretation.}
The metric measures full-result availability at detected lexical
reuse events on these fixed traces. Lexical repetition does not
establish that a later operation depended on the earlier result,
and exact matching can miss reuse expressed differently. Higher
recall therefore indicates better retention for this reference event
set; it does not independently establish the precision or recall of
TFA's dependency detection or the causal correctness of its edges.

The recorded AppWorld histories contain 3,174 result-level reuse events.
Table~\ref{tab:appworld-reuse-events} reports their distribution by
result age.

\begin{table}[t]
\centering
\small
\setlength{\tabcolsep}{6pt}
\renewcommand{\arraystretch}{1.05}
\caption{AppWorld reuse events by result age.}
\label{tab:appworld-reuse-events}
\begin{tabular}{lr}
\toprule
\textbf{Result age at reuse} & \textbf{Events} \\
\midrule
$\leq 3$ turns & 713 \\
4--10 turns    & 1,124 \\
11--25 turns   & 1,099 \\
$>25$ turns    & 238 \\
\midrule
Total          & 3,174 \\
\bottomrule
\end{tabular}
\end{table}

The recorded 8-objective QA histories contain 335 result-level reuse
events across 100 tasks, using one GPT-4.1 trace per task with
\textit{Full history}. We use the same event-counting convention as
for AppWorld. Table~\ref{tab:qa-reuse-events} reports the corresponding
distribution by result age.

\begin{table}[t]
\centering
\small
\setlength{\tabcolsep}{6pt}
\renewcommand{\arraystretch}{1.05}
\caption{8-objective QA reuse events by result age.}
\label{tab:qa-reuse-events}
\begin{tabular}{lr}
\toprule
\textbf{Result age at reuse} & \textbf{Events} \\
\midrule
$\leq 3$ turns & 224 \\
4--10 turns    & 70 \\
11--25 turns   & 41 \\
$>25$ turns    & 0 \\
\midrule
Total          & 335 \\
\bottomrule
\end{tabular}
\end{table}

\section{Inference Cost Accounting and Numerical Results}
\label{app:cost-results}

\subsection{Cost Comparison Tasks}
\label{app:cost-cohort}

For each model--benchmark setting, we identify cost-comparison
tasks using reference runs without context management. A task
is selected if its accumulated execution history exceeds
$B=6$K tokens before a model invocation, using the history-token
accounting in Appendix~\ref{app:budget-accounting}. Let
$\mathcal{T}_{\mathrm{CM}}$ denote this fixed set of task IDs.
We use this set to compare mean inference cost across all methods,
including \textit{Full history}. Cost averages include both
successful and unsuccessful runs.

\subsection{Token Usage, Standardization, and Normalization}
\label{app:cost-standardization}

\paragraph{Token usage.}
For method $m$ and task $i$, we count input tokens $P_{m,i}$,
output tokens $O_{m,i}$, and embedding tokens $E_{m,i}$ across
all agent and context-management calls. Compression and
summarization calls are included, including oversized-result
compaction if triggered; $E_{m,i}=0$ when no embedding calls are used.
Input-token accounting includes the complete model input, including
prompt content outside the history budget.
The history-token count used to enforce $B$ and the token usage
used to calculate inference cost therefore have different scopes.

\paragraph{Standardized cost.}
We compute each task's cost $C_{m,i}$ using
Equation~\ref{eq:standardized-inference-cost}.
Within each execution model and benchmark, $\rho$ is the cache
fraction observed in uncapped \textit{Full history}; the same value
is used to price every method. It is not re-estimated from each
method's own cache hits and can differ across model--benchmark
settings. This standardization compares token costs under common
prices and cache assumptions. It does not estimate a provider bill
or assert identical cache rates in deployment. Graph storage and
local selection computation are outside this token-cost measure.

\paragraph{Averaging and normalization.}
For each model--benchmark setting, mean and normalized inference
cost are
\begin{equation}
\overline{C}_m
=\frac{1}{|\mathcal{T}_{\mathrm{CM}}|}
 \sum_{i\in\mathcal{T}_{\mathrm{CM}}}C_{m,i},
\qquad
\widetilde{C}_m
=\frac{\overline{C}_m}
       {\overline{C}_{\mathrm{Full\ history}}}.
\label{eq:app-normalized-subset-cost}
\end{equation}
Figure~\ref{fig:rq3-cost} plots $\widetilde{C}_m$, with
\textit{Full history} normalized to 1 in each setting. This is a
ratio of mean costs, not a mean of task-level cost ratios.

\subsection{Numerical Performance and Cost Results}
\label{app:performance-cost-values}

Tables~\ref{tab:app-cost-gpt41}--\ref{tab:app-cost-deepseek} provide
the numerical values underlying Figure~\ref{fig:rq3-cost} and
Table~\ref{tab:relative-performance-cost}. They report overall
AppWorld TGC or QA F1, mean inference cost $\overline{C}_m$, and
normalized cost $\widetilde{C}_m$ using the accounting above.
AppWorld TGC uses all 168 Test-Normal tasks and all 417
Test-Challenge tasks; percentages are rounded to one decimal place.

The relative changes in Table~\ref{tab:relative-performance-cost}
use each comparison method's value as the denominator. Performance
changes are relative percentages; negative cost changes indicate
cost reductions.

\begin{table*}[t]
\caption{Overall task performance and mean inference cost for GPT-4.1.
Performance is measured across all tasks; cost is measured on the same
tasks that trigger context management when accumulated execution
history exceeds $B=6$K. Cost is in reference USD; normalized cost
uses uncapped \textit{Full history} ($=1$) on the same tasks in each
benchmark.}
\label{tab:app-cost-gpt41}
\centering
\footnotesize
\setlength{\tabcolsep}{2.5pt}
\renewcommand{\arraystretch}{1.08}
\resizebox{\textwidth}{!}{%
\begin{tabular}{@{}lrrrrrrrrr@{}}
\toprule
\multirow{2}{*}{\textbf{Method}}
& \multicolumn{3}{c}{\textbf{AppWorld Test-Normal}}
& \multicolumn{3}{c}{\textbf{AppWorld Test-Challenge}}
& \multicolumn{3}{c}{\textbf{8-objective QA}} \\
\cmidrule(lr){2-4}
\cmidrule(lr){5-7}
\cmidrule(lr){8-10}
& \textbf{TGC (\%)} & \textbf{Cost} & \textbf{Norm.}
& \textbf{TGC (\%)} & \textbf{Cost} & \textbf{Norm.}
& \textbf{F1 (\%)} & \textbf{Cost} & \textbf{Norm.} \\
\midrule
\textit{Full history}
& 77.4 & 0.177 & 1.000 & 51.1 & 0.302 & 1.000 & 50.8 & 0.167 & 1.000 \\

Prune
& 66.1 & 0.124 & 0.701 & 46.3 & 0.221 & 0.732 & 51.2 & 0.153 & 0.916 \\

Compact (LLM)
& 67.3 & 0.121 & 0.684 & 49.4 & 0.219 & 0.725 & 52.0 & 0.157 & 0.940 \\

Recency + Summary
& 70.2 & 0.122 & 0.689 & 49.6 & 0.258 & 0.854 & 48.5 & 0.150 & 0.898 \\

Semantic Retrieval
& 70.2 & 0.147 & 0.831 & 48.4 & 0.238 & 0.788 & 46.4 & 0.150 & 0.898 \\

ACON
& 73.2 & 0.141 & 0.797 & 49.9 & 0.234 & 0.775 & 48.1 & 0.159 & 0.952 \\

\rowcolor{crblue}
\textbf{\sysname{}}
& 75.6 & 0.120 & 0.678 & 55.4 & 0.215 & 0.712 & 53.1 & 0.150 & 0.898 \\

\bottomrule
\end{tabular}%
}
\end{table*}

\begin{table*}[t]
\caption{Overall task performance and mean inference cost for Muse Glimmer 30B.
Performance is measured across all tasks; cost is measured on the same
tasks that trigger context management when accumulated execution
history exceeds $B=6$K. Cost is in reference USD; normalized cost
uses uncapped \textit{Full history} ($=1$) on the same tasks in each
benchmark.}
\label{tab:app-cost-muse}
\centering
\footnotesize
\setlength{\tabcolsep}{2.5pt}
\renewcommand{\arraystretch}{1.08}
\resizebox{\textwidth}{!}{%
\begin{tabular}{@{}lrrrrrrrrr@{}}
\toprule
\multirow{2}{*}{\textbf{Method}}
& \multicolumn{3}{c}{\textbf{AppWorld Test-Normal}}
& \multicolumn{3}{c}{\textbf{AppWorld Test-Challenge}}
& \multicolumn{3}{c}{\textbf{8-objective QA}} \\
\cmidrule(lr){2-4}
\cmidrule(lr){5-7}
\cmidrule(lr){8-10}
& \textbf{TGC (\%)} & \textbf{Cost} & \textbf{Norm.}
& \textbf{TGC (\%)} & \textbf{Cost} & \textbf{Norm.}
& \textbf{F1 (\%)} & \textbf{Cost} & \textbf{Norm.} \\
\midrule
\textit{Full history}
& 81.5 & 0.137 & 1.000 & 58.3 & 0.161 & 1.000 & 50.4 & 0.269 & 1.000 \\

Prune
& 75.6 & 0.118 & 0.861 & 51.3 & 0.139 & 0.863 & 49.3 & 0.243 & 0.903 \\

Compact (LLM)
& 80.4 & 0.119 & 0.869 & 54.9 & 0.155 & 0.963 & 50.4 & 0.259 & 0.963 \\

Recency + Summary
& 79.8 & 0.124 & 0.905 & 54.4 & 0.132 & 0.820 & 48.2 & 0.245 & 0.911 \\

Semantic Retrieval
& 78.0 & 0.135 & 0.985 & 52.8 & 0.141 & 0.876 & 51.1 & 0.235 & 0.874 \\

ACON
& 78.6 & 0.123 & 0.898 & 55.4 & 0.147 & 0.913 & 51.8 & 0.252 & 0.937 \\

\rowcolor{crblue}
\textbf{\sysname{}}
& 86.3 & 0.120 & 0.876 & 60.0 & 0.129 & 0.801 & 52.7 & 0.223 & 0.829 \\

\bottomrule
\end{tabular}%
}
\end{table*}

\begin{table*}[t]
\caption{Overall task performance and mean inference cost for DeepSeek V4 Pro.
Performance is measured across all tasks; cost is measured on the same
tasks that trigger context management when accumulated execution
history exceeds $B=6$K. Cost is in reference USD; normalized cost
uses uncapped \textit{Full history} ($=1$) on the same tasks in each
benchmark.}
\label{tab:app-cost-deepseek}
\centering
\footnotesize
\setlength{\tabcolsep}{2.5pt}
\renewcommand{\arraystretch}{1.08}
\resizebox{\textwidth}{!}{%
\begin{tabular}{@{}lrrrrrrrrr@{}}
\toprule
\multirow{2}{*}{\textbf{Method}}
& \multicolumn{3}{c}{\textbf{AppWorld Test-Normal}}
& \multicolumn{3}{c}{\textbf{AppWorld Test-Challenge}}
& \multicolumn{3}{c}{\textbf{8-objective QA}} \\
\cmidrule(lr){2-4}
\cmidrule(lr){5-7}
\cmidrule(lr){8-10}
& \textbf{TGC (\%)} & \textbf{Cost} & \textbf{Norm.}
& \textbf{TGC (\%)} & \textbf{Cost} & \textbf{Norm.}
& \textbf{F1 (\%)} & \textbf{Cost} & \textbf{Norm.} \\
\midrule
\textit{Full history}
& 82.7 & 0.1425 & 1.000 & 77.0 & 0.236 & 1.000 & 56.5 & 0.476 & 1.000 \\

Prune
& 81.0 & 0.1185 & 0.832 & 57.6 & 0.180 & 0.763 & 55.2 & 0.424 & 0.891 \\

Compact (LLM)
& 82.1 & 0.126 & 0.884 & 70.7 & 0.222 & 0.941 & 58.2 & 0.444 & 0.933 \\

Recency + Summary
& 82.1 & 0.117 & 0.821 & 71.2 & 0.179 & 0.758 & 58.5 & 0.435 & 0.914 \\

Semantic Retrieval
& 78.0 & 0.120 & 0.842 & 67.1 & 0.162 & 0.686 & 55.3 & 0.419 & 0.880 \\

ACON
& 82.1 & 0.1305 & 0.916 & 67.9 & 0.212 & 0.898 & 57.4 & 0.456 & 0.958 \\

\rowcolor{crblue}
\textbf{\sysname{}}
& 85.7 & 0.1155 & 0.811 & 76.3 & 0.181 & 0.767 & 60.5 & 0.425 & 0.893 \\

\bottomrule
\end{tabular}%
}
\end{table*}

\section{Runtime Adapters}
\label{app:adapters}

\sysname{} runs as a separate server. A runtime adapter connects an
agent runtime to the server by intercepting the model context before
each model call. Table~\ref{tab:adapters} summarizes five runtimes
and the integration point used for each.

The runtimes expose this integration point in three ways. First,
opencode provides a hook that runs before the model request is
constructed. Second, the AppWorld ReAct agent and ACON UnifiedAgent
expose methods or properties that construct the conversation history,
which the adapter overrides. Third, OpenClaw and the Meta ARE agent
allow the model endpoint to be replaced with an OpenAI-compatible
proxy, which applies \sysname{} before forwarding the request.

These integrations demonstrate that \sysname{} can connect to runtimes
that expose a point where the outgoing model context can be inspected
and modified, either directly or through a configurable model endpoint.
We do not claim compatibility with every agent runtime. A runtime that
exposes neither mechanism cannot use the current adapter design.

\begin{table*}[t]
\centering
\footnotesize
\setlength{\tabcolsep}{4pt}
\caption{Agent runtimes connected to \sysname{}. For each runtime, we
list the adapter insertion point, the runtime component modified by the
integration, and its use in this work. Prompting, tool execution,
parsing, and grading are not rewritten.}
\label{tab:adapters}

\begin{tabular}{@{}p{1.05in}p{1.6in}p{1.6in}p{0.8in}@{}}
\toprule
Agent runtime & Insertion point & Integration changes & Used for \\
\midrule

opencode &
\texttt{experimental.chat.\allowbreak messages.transform} plugin hook;
the plugin is specified in \texttt{opencode.json} &
The runtime's own compaction and pruning are disabled so that only
\sysname{} manages historical context. &
Integration only \\

\addlinespace

AppWorld ReAct agent &
\texttt{trimmed\_messages} property replaced by a mixin on the live
agent object &
One property. Prompting, parsing, retry logic, tool use, runner,
configuration, and CLI are unchanged. &
AppWorld \\

\addlinespace

ACON UnifiedAgent (smolagents) &
\texttt{MemoryManager} subclass overriding
\texttt{get\_conversation\_history}; the class reference is rebound
before agent construction &
Memory class only. Runner, agent, prompts, retriever, and grading are
unchanged. &
8-objective QA \\

\addlinespace

OpenClaw (CLI in Docker) &
OpenAI-compatible proxy registered as a custom provider in
\texttt{openclaw.json} &
No runtime code is modified; context rendering is applied by the proxy
before the request is forwarded. &
Sanity check \\

\addlinespace

Meta ARE default agent (GAIA2) &
The same proxy, selected with the
\texttt{--provider local --endpoint} flags &
No runtime code is modified. &
Integration check \\

\bottomrule
\end{tabular}
\end{table*}

\end{document}